\documentclass[conference]{IEEEtran}
\usepackage{amsmath,cite,nicefrac,siunitx,array,framed,booktabs,color,float,epsfig,wrapfig,graphics,graphicx,subcaption,textcomp,amssymb,setspace,latexsym,fancyhdr,url,enumerate,algpseudocode,graphics,xparse,xspace,multirow,csvsimple,balance,hyperref,tikz,tabularx}
\usepackage[ruled]{algorithm2e}
\usetikzlibrary{calc}

\newcommand{\DeepSense}{\textsc{DeepSense}}
\newcommand{\SSVax}{\textsc{SSVax}}
\newcommand{\SemiSS}{\textsc{SemiSS}}
\newcommand{\FLTrust}{\textsc{FLTrust}}
\newcommand{\SemiFL}{\textsc{SemiFL}}
\newcommand{\Median}{\textsc{Median}}
\newcommand{\TrMean}{\textsc{TrMean}}
\newcommand{\FLShield}{\textsc{FLShield}}
\newcommand{\Krum}{\textsc{Krum}}
\newcommand{\Kmeans}{\textsc{k-means}}
\newcommand{\smallsize}{\small}

\begin{document}
\title{Self-Adaptive and Robust Federated Spectrum Sensing without Benign Majority for Cellular Networks}

\newif\ifanonymous

\ifanonymous
\author{\IEEEauthorblockN{Anonymous Authors}}
\else
\author{\IEEEauthorblockN{Ngoc Duy Pham\\Thusitha Dayaratne\\Carsten Rudolph}
	\IEEEauthorblockA{Monash University}
	\and
	\IEEEauthorblockN{Viet Vo}
	\IEEEauthorblockA{Swinburne University\\of Technology}
        \and
        \IEEEauthorblockN{Shangqi Lai\\Sharif Abuadbba\\Hajime Suzuki}
        \IEEEauthorblockA{CSIRO's Data61}
        \and
        \IEEEauthorblockN{Xingliang Yuan}
        \IEEEauthorblockA{The University\\of Melbourne}}
	
\IEEEoverridecommandlockouts
\makeatletter\def\@IEEEpubidpullup{6.5\baselineskip}\makeatother
\fi

\maketitle

\begin{abstract}
Advancements in wireless and mobile technologies, including 5G advanced and the envisioned 6G, are driving exponential growth in wireless devices. However, this rapid expansion exacerbates spectrum scarcity, posing a critical challenge. Dynamic spectrum allocation (DSA)--which relies on sensing and dynamically sharing spectrum--has emerged as an essential solution to address this issue. While machine learning (ML) models hold significant potential for improving spectrum sensing, their adoption in centralized ML-based DSA systems is limited by privacy concerns, bandwidth constraints, and regulatory challenges. To overcome these limitations, distributed ML-based approaches such as Federated Learning (FL) offer promising alternatives. This work addresses two key challenges in FL-based spectrum sensing (FLSS). First, the scarcity of labeled data for training FL models in practical spectrum sensing scenarios is tackled with a semi-supervised FL approach, combined with energy detection, enabling model training on unlabeled datasets. Second, we examine the security vulnerabilities of FLSS, focusing on the impact of data poisoning attacks. Our analysis highlights the shortcomings of existing majority-based defenses in countering such attacks. To address these vulnerabilities, we propose a novel defense mechanism inspired by vaccination, which effectively mitigates data poisoning attacks without relying on majority-based assumptions. Extensive experiments on both synthetic and real-world datasets validate our solutions, demonstrating that FLSS can achieve near-perfect accuracy on unlabeled datasets and maintain Byzantine robustness against both targeted and untargeted data poisoning attacks, even when a significant proportion of participants are malicious.
\end{abstract}


\IEEEpeerreviewmaketitle

\section{Introduction}\label{introduction}
The rapid expansion of wireless communication in recent years has been driven by its convenience, ubiquity, cost-effectiveness, and technological advancements. This growth is expected to accelerate further in the beyond 5G (B5G) and 6G era, where the convergence of physical and digital domains will enable transformative applications like smart cities and the Internet of Everything \cite{6GandBeyond20Akyildiz,SecurityPrivacy6G23Porambage}. These emerging applications require substantial bandwidth. While millimeter-wave frequencies (30GHz -- 300GHz) play a central role in 5G, and sub-THz and THz bands (100GHz -- 1THz) are anticipated to be critical for 6G, the mid-band spectrum (1GHz -- 6GHz) is expected to remain heavily utilized due to its optimal balance between coverage and capacity \cite{MillimeterWave22Moltchanov}. Consequently, efficient spectrum sharing in the licensed mid-band is essential to support current and future wireless technologies, including WiFi, LTE, 5G, and 6G networks.

\begin{figure}[h]
\centering
\includegraphics[width=.85\linewidth]{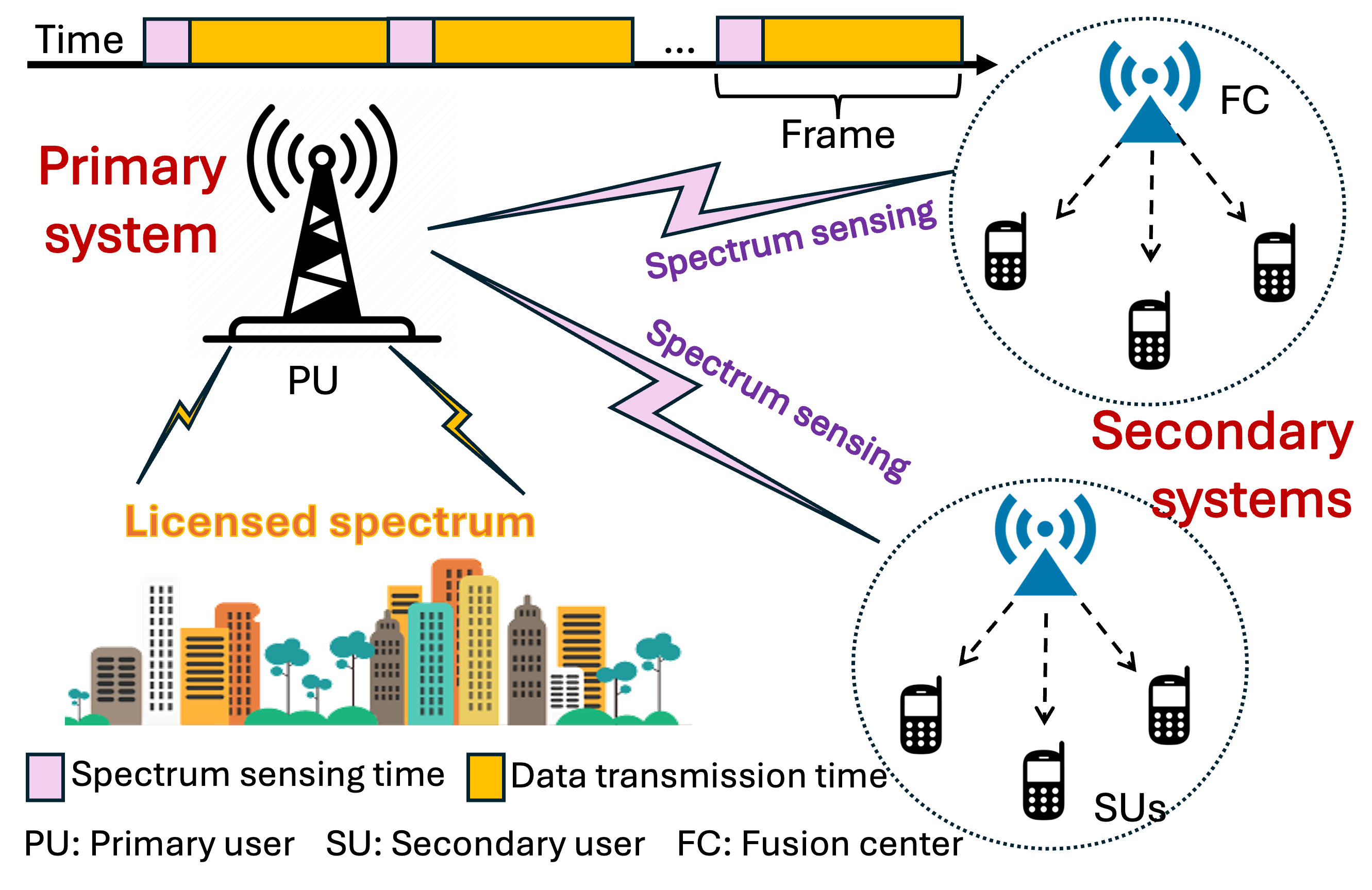}
\caption{Spectrum reuse in cognitive radio networks, enabling secondary systems to access the spectrum of primary systems dynamically.}
\label{fig:spectrum-sensing}
\end{figure}

Spectrum sensing (SS) is essential for identifying spectrum gaps, or `holes', that secondary/unlicensed users (SUs) can opportunistically exploit without interfering with primary/licensed users (PUs). As illustrated in Fig.~\ref{fig:spectrum-sensing}, PUs do not always utilize their allocated spectrum continuously, creating opportunities for SUs to sense unused spectrum and transmit data during periods of PU inactivity. While SUs can perform autonomous sensing, this approach is often unreliable due to wireless propagation effects \cite{SecureFL23Wasilewska}. Cooperative sensing addresses these challenges by aggregating sensing data from multiple devices at a fusion center (FC), which determines spectrum occupancy--typically using majority voting. However, traditional sensing methods struggle to effectively leverage the inherent time, frequency, and spatial correlations in wireless signals, limiting their performance \cite{SecureFL23Wasilewska}. Machine learning (ML) offers a promising alternative by extracting complex and nuanced features from the data, enabling significantly improved sensing accuracy compared to conventional approaches~\cite{SpectrumSensing12Axell}.

ML algorithms have been widely applied to SS, addressing tasks such as spectrum classification \cite{ML4LTE19Wasilewska,MLbasedSS15Xue,SSbasedSoftmax16Zhang} and cooperative sensing via data fusion \cite{ML2DataFusion14Mikaeil,RLbasedCS10Lo,Bayesian4CSS17Tavares,MaliciousDetection21Hossain}. Recently, the authors in \cite{DeepSense21Uvaydov} proposed \DeepSense, a deep learning-based approach that achieved near-perfect accuracy in SS across multiple channels. However, this method requires transmitting raw training data to the FC, which imposes significant communication overhead, increases latency, and raises privacy concerns. Centralized data aggregation exposes sensitive information in user equipment (UE) sensing data, such as historical location or device identity, leading to compliance risks under regulations like the General Data Protection Regulation (GDPR) \cite{GDPR2016EU}.

To tackle these limitations, federated learning (FL) has emerged as a promising approach for SS \cite{FL45G22Wasilewska,FL4CR22Chen,SecureFL23Wasilewska}. FL enables SUs to collaboratively train ML models by exchanging model parameters instead of raw data, thereby reducing bandwidth consumption and preserving data privacy. Lightweight ML models, such as \DeepSense, which has approximately $16k$ parameters, demonstrate the feasibility of deploying efficient solutions on mobile devices \cite{DeepSense21Uvaydov}, positioning FL as a critical enabler for AI-driven 6G networks. Despite its potential, significant advancements are necessary to make FL-based SS (FLSS) practical and reliable for efficient spectrum utilization.

In this work, we explore the application of FLSS in current and future wireless networks, addressing key challenges and proposing innovative solutions to enhance both performance and security. One of the primary challenges we tackle is the limited availability of labeled data for FL training--a critical issue often overlooked or assumed to be resolved through manual data processing in prior studies \cite{DeepSense21Uvaydov,DeepSweep24Robinson,FL4CR22Chen,LowSS23Mei,LabelFree24Milosheski}. For instance, datasets used in \cite{DeepSense21Uvaydov} contain hundreds of thousands of transmissions. While SUs can naturally sense the spectrum, labeling such large datasets is prohibitive expensive, emphasizing the importance of leveraging ML on unlabeled data. Inspired by semi-supervised learning techniques \cite{TriTraining05Zhou,PseudoLabel13Lee}, we propose \SemiSS, a novel method that combines semi-supervised FL \cite{SemiFL24Diao} with energy detection \cite{SpectrumSensing09Yucek}. This approach enhances model training on unlabeled data by utilizing a small, centrally labeled dataset. Energy detection facilitates self-adaptive label correction at the SUs, enabling the system to achieve SS accuracy comparable to that of fully labeled datasets, significantly reducing the reliance on extensive manual labeling.

On the security front, FLSS inherits vulnerabilities from traditional FL, particularly susceptibility to poisoning attacks. The distributed and often anonymous nature of FLSS, especially in emerging open radio networks \cite{5GinaBox}, amplifies the risk of adversarial SUs compromising the global model, leading to degraded performance and potential network collisions. We analyze the impact of poisoning attacks on FLSS model training and identify weaknesses in existing defense mechanisms, which often rely on majority-based assumptions \cite{MESAS23Kraub,FLShield24Kabir,ManipulateByzantine21Shejwalkar,MultiKrum17Blanchard,TrimmedMean18Yin,FangAttack20Fang}. These mechanisms are often ineffective when malicious clients form a significant portion of participants, allowing them to evade detection and corrupt the model.
In cross-device FLSS \cite{FLSurvey23Li} with large participant pools, such as radio networks, a minority of malicious clients can dominate specific training rounds, undermining state-of-the-art majority-based defenses.
To address these limitations, we propose \SSVax, a novel defense mechanism inspired by the concept of vaccination. Unlike traditional approaches, \SSVax\ avoids majority-based assumptions. It strengthens FLSS by using pseudo-attacks to generate `vaccines' that detect and filter malicious model updates, ensuring secure and robust model aggregation. In summary, the key contributions of this paper are as follows:

\begin{itemize}
\item We propose \SemiSS, which leverages semi-supervised learning and energy detection to locally label unlabeled data, effectively addressing the challenge of limited labeled data in FLSS applications. \SemiSS\ achieves performance comparable to fully supervised learning and accelerates convergence by leveraging frequency-domain spectrum data.

\item We analyze the impact of data poisoning attacks on FLSS and critically evaluate existing defense mechanisms. Highlighting the significant limitations of majority-based defenses, we propose \SSVax, a vaccination-inspired defense mechanism designed to secure FLSS systems against data poisoning attacks without relying on majority-based assumptions.

\item We validate our proposed methods through extensive evaluations on synthetic and real-world datasets. Results demonstrate the effectiveness of FLSS, showing strong performance on unlabeled data and robust resilience against both targeted and untargeted attacks, even at high poisoning rates. 
\end{itemize}


\section{Background and Related Works}\label{background}
\subsection{Spectrum sensing}
In telecommunications, the spectrum refers to the range of electromagnetic frequencies used for wireless communication. This range includes radio frequencies utilized in various applications, such as radio broadcasting, television, Wi-Fi, cellular networks, and satellite systems. The spectrum is divided into bands, each allocated for specific communication purposes, with frequencies typically measured in hertz (Hz) \cite{5GB22Lin}. Within these bands, channels represent narrower frequency ranges used for individual communication links. Wireless communication relies on the transmission and reception of raw data via antennas, which capture electromagnetic waves and convert them into electrical signals. These signals are digitized into a series of data points that represent the amplitude and phase of the original signal, commonly referred to as in-phase and quadrature (IQ) samples. IQ samples, which capture the signal in the time domain, can be transformed into the frequency domain using a Fourier transform \cite{FourierTransform03Cooley}. This frequency-domain representation reveals the signal's frequency components, facilitating spectrum analysis for detecting patterns, identifying interference, and enabling efficient spectrum utilization \cite{SpectrumSensing09Yucek}.

Spectrum sensing is a critical process in wireless communication, aimed at optimizing the utilization of the radio frequency spectrum. By continuously monitoring the spectrum, SS systems detect the presence of licensed users (PUs) and identify unused frequency bands or channels. This enables SUs (unlicensed) to opportunistically access the spectrum without causing interference to PUs when it is idle \cite{SpectrumSensing12Axell}. The received (sensed) signal at the SU is used to decide between two states: $\mathcal{H}_0$, where the PU signal is absent, or $\mathcal{H}_1$, where the PU signal is present, as defined below:
\begin{equation}
\begin{split}
\mathcal{H}_0&:y(t)=w(t)\\
\mathcal{H}_1&:y(t)=x(t)+w(t),
\end{split}
\end{equation}

\noindent where $y(t)$ represents the received signal, $x(t)$ denotes the PU's transmitted signal, and $w(t)$ is the noise affecting the transmitted signal at time $t$. Effective SS is essential to address spectrum scarcity and enhance spectrum utilization. Energy detection is one of the most fundamental SS techniques that measures the energy received over a finite time interval and compares it to a predetermined threshold. Given a number $N$ of sampled signals, the energy detection method computes the total energy detected (TED) using the following formula:
\begin{equation}
\text{TED} = \sum_{t=1}^N \left[y(t)\right]^2 \ \ \ \
\begin{cases}
<\lambda: & \text{PU signal absent}\\
\geq\lambda: & \text{PU signal present}.
\end{cases}
\end{equation}

\noindent If the energy exceeds the threshold $\lambda$, the PU is assumed to be active; otherwise, it is considered inactive. However, the effectiveness of energy detection heavily depends on the choice of the threshold value. An inappropriate threshold can lead to suboptimal performance, resulting in either missed detections or false alarms. In \cite{DeepSense21Uvaydov}, energy detection achieved approximately $60\%$ accuracy, whereas ML-based approaches demonstrated near-perfect results, representing the state-of-the-art (SOTA). This substantial improvement is attributed to ML's ability to learn complex system dynamics directly from raw data, circumventing the limitations of traditional SS methods that depend on predefined mathematical models \cite{DeepSweep24Robinson}.

\subsection{Federated learning for SS}
Federated learning \cite{FL16McMahan,FL17McMahan,FL19Sheller} is a distributed machine learning paradigm that allows $\mathbf{N}$ clients and a central aggregator to collaboratively build a global model $G$ without sharing raw data. Recently, there has been growing interest in applying FL to SS \cite{SecureFL23Wasilewska}, leveraging its distributed architecture and privacy-preserving design principles. In an FL-based SS system, SUs utilize commercial off-the-shelf hardware to train local SS models on their own sensing data. Rather than transmitting raw data to a central FC, only model parameters (weights) $W$ are shared and aggregated at the FC to build a global model. This approach ensures local data privacy while reducing communication overhead. Additionally, SUs can perform local SS tasks for faster decision-making, which enables efficient spectrum utilization to meet communication demands. The basic training procedure for an FLSS system is outlined in Alg. \ref{alg:FLbasic} with the relevant notations summarized in Table \ref{tab:notations}. In this FLSS approach, having a labeled training dataset is crucial \cite{FL4CR22Chen}. However, obtaining such labels remains a significant challenge, as existing literature often assumes that labeled data is readily available or relies on labor-intensive manual labeling processes.

\begin{table}[t]\centering\smallsize
\caption{Summary of key notations.}\label{tab:notations}
\begin{tabularx}{\linewidth}{|l|X|}\hline
\textbf{Notation} & \textbf{Meaning} \\\hline\hline
$\mathcal{D}^S$, $\mathcal{D}^U$ & Labeled and unlabeled datasets \\\hline
$W_G$, $W_{FC}$, $W_{SU}$ & Global, FC, and SU model parameters \\\hline
$\mathbf{T}$, $\mathbf{E}$ & Number of FL training rounds and local training epochs \\\hline
$\mathbf{R}_S$, $\mathbf{R}_M$, $\mathbf{R}_C$ & Ratios of selected SUs, malicious SUs, and label correction \\\hline
$\mathbf{N}$, $\mathbf{N}_S$ & Total number of SUs and number of SUs selected per FL training round \\\hline
\end{tabularx}
\end{table}

\begin{algorithm}[t]\smallsize
\caption{FLSS model training procedure.}\label{alg:FLbasic}
\SetAlgoLined\SetKwProg{Fn}{}{:}{}
\Fn{System execution} {
    \nl FC initializes a global model $W_G^0$\\
    \nl\For{each training round $t = 1,2,\ldots\mathbf{T}$} {
        \nl$\mathcal{S}^t\gets$ uniformly sample $\mathbf{N}_S$ SUs\\
        \nl\For{each SU$_i\in\mathcal{S}^t$ \textbf{in parallel}} {
            \nl$W_{SU_i}^t\gets \textsc{SU\_Update}(W_G^{t-1})$
        }
        \nl$W_G^{t}\gets 1/\mathbf{N}_S\sum_{SU_i\in\mathcal{S}^t}W_{SU_i}^{t}$
    }
    \nl\Return $W_G^\mathbf{T}$\\
}
\Fn{$\textsc{SU\_Update}(W_G)$} {
    \nl$W_{SU}\gets W_G$\\
    \nl\For{each local epoch from $1$ to $\mathbf{E}$} {
        \nl\For{each local data batch $(x,y)$} {
            \nl$L\gets\mathcal{L}(f(x,W_{SU}),y)$\\
            \nl$W_{SU}\gets W_{SU}-\eta\triangledown_WL$
        }
    }
    \nl\Return $W_{SU}$ to FC
}
\end{algorithm}

\subsection{Semi-supervised learning}
Semi-supervised learning (SSL) \cite{TriTraining05Zhou} is an established technique for addressing the scarcity of labeled data. SSL typically involves training a supervised model on a small set of labeled data, which is then used to generate pseudo-labels for the unlabeled data \cite{PseudoLabel13Lee}. These pseudo-labeled samples are combined with the labeled data to retrain the model, enhancing its overall performance. However, the effectiveness of SSL models is highly contingent on the quality of the pseudo-labels, as inaccurate labels can hinder learning outcomes \cite{SemiFL24Diao}. To address this challenge, we propose a pseudo-label correction mechanism that utilizes energy detection to improve the reliability of pseudo-labels in FLSS applications.

\subsection{Poisoning attacks on FLSS and defenses}
In addition to the challenge of obtaining labeled data, FLSS also inherits security vulnerabilities from FL, particularly its susceptibility to poisoning attacks. The anonymous and distributed nature of FL exposes it to adversaries who may compromise participants and manipulate the aggregation process \cite{ManipulateByzantine21Shejwalkar}. By controlling a subset of participants, attackers can launch poisoning attacks with various malicious objectives, such as degrading model accuracy, causing interference with PUs, or manipulating spectrum occupancy data to monopolize a channel. Defenses against poisoning attacks in FL typically include robust aggregation techniques or detect-and-filter mechanisms. Many existing defenses rely on the \emph{majority rule}, which assumes that the majority of model updates are benign. While this approach is effective in certain contexts, it becomes vulnerable in cross-device FL systems \cite{FLSurvey23Li}, where malicious updates may outnumber benign ones, undermining the robustness of the aggregation process.

Popular defense mechanisms for ensuring Byzantine-robust FL are typically deployed at the aggregation server. \Median~\cite{TrimmedMean18Yin} is a coordinate-wise aggregation rule that sorts the values of each parameter across all local models and selects the median value for the global model. Similarly, \TrMean~\cite{TrimmedMean18Yin} removes the largest and smallest $\mathbf{N}_M$ values, where $\mathbf{N}_M$ represents the number of assumed malicious clients (with the constraint $\mathbf{N}_M<\mathbf{N}/2$), and aggregates the remaining values into the global model. Leveraging the intuition that malicious models deviate significantly from benign ones, \Krum~\cite{MultiKrum17Blanchard} selects local models that are closest (i.e., with the smallest Euclidean distance) to their $\mathbf{N}-\mathbf{N}_M-2$ neighbors for global aggregation. \FLShield, a validation-based defense approach, validates local models on client-held datasets prior to aggregation, under the assumption that malicious clients constitute less than $50\%$ of the total clients. In a similar vein, MESAS \cite{MESAS23Kraub} clusters local models into two fixed groups, assuming that the smaller group contains the malicious models ($\mathbf{N}_M<\mathbf{N}/2)$. Malicious detection is performed in a cascaded manner using multiple carefully chosen metrics to counter adaptive attackers \cite{AdversarialAdaption24KrauB}. Unlike methods relying on the majority rule, \FLTrust~\cite{FLTrust22Cao} uses a server-side validation dataset to train a server model and calculate trust scores for local models based on their cosine similarity to the server model. These trust scores are then used as weights for model aggregation. However, our analysis reveals that this method may still assign non-zero weights to malicious models, leaving potential vulnerabilities.

\section{Problem Statement}\label{problem}
Despite the significant potential of the B5G era, challenges such the scarcity of labeled training data and inherent security threats, including data poisoning attacks, hinder the realization of practical FLSS systems. This section outlines the assumptions underlying our proposed solutions, as well as the threat model used to analyze poisoning attacks and the corresponding defenses.

\subsection{System model}
In a supervised SS classification task, the dataset is represented as $\mathcal{D}=\{x_i,y_i\}$, where $x_i$ denotes the input features, and $y_i=\{0,1\}$ is the corresponding class label. For a \emph{semi}-supervised classification task, two datasets are considered: a labeled dataset $\mathcal{D}^S=\{x^S_i,y^S_i\}$ and an unlabeled dataset $\mathcal{D}^U=\{x^U_i\}$.

This study focuses on semi-supervised FLSS, where $\mathcal{D}^U$ is uniformly distributed across $\mathbf{N}$ number of SUs (clients), while a small labeled dataset $\mathcal{D}^S$ resides on the FC (central server), with $|\mathcal{D}^S|\ll|\mathcal{D}^U|$. The learning framework involves three parameterized models: the global model $W_G$, the FC model $W_{FC}$, and the SU model $W_{SU}$. These models share the same architecture, $f:(x,W)\mapsto~f(x,W)$, which maps input $x$ and model parameters $W$ to model outputs. A standard supervised loss function $\mathcal{L(\cdot)}$ quantifies the difference between the model outputs and their corresponding labels.

The training process follows a typical cross-device FL approach \cite{FLSurvey23Li}. Over $\mathbf{T}$ training rounds, the FC randomly selects $\mathbf{N}_S$ SUs to perform local training for $\mathbf{E}$ epochs. The locally trained models, referred to as updates, are then aggregated at the FC using the standard \textsc{FedAvg} method \cite{FL17McMahan}. Table \ref{tab:notations} summarizes the key notations used throughout this paper.

\subsection{Threat model}
Consistent with the poisoning attack threat model described in the literature \cite{FLSurvey24Lyu}, we consider adversaries controlling multiple malicious SUs, which may be fake or compromised legitimate devices, while the FC remains uncompromised. This scenario is realistic in radio networks, where SUs (e.g., mobile or IoT devices) collaborating in FLSS for spectrum sharing \cite{5GinaBox} are more susceptible to compromise. Conversely, compromising the FC is less feasible, as it is typically deployed in the control plane, protected by robust security measures.
Adversaries aim to: (1) degrading the global model's detection accuracy, causing interference or collisions between PUs and SUs, and (2) pursuing selfish objectives, such as falsely reporting channels as always busy to block others or always available to cause collisions.

To achieve these goals, the adversary launches data poisoning attacks that are either \emph{untargeted} (disrupting all channels) or \emph{targeted} (focused on specific channels). The success of an attack is measured by the attack success rate (ASR), defined as the rate of successful misclassifications to the target class:
\begin{equation}
    \text{ASR} = \mathbb{E}_{(x,y)\sim\mathcal{D},\,y=y_s} \left[\Pr(f(x))=y_t)\right],
\end{equation}

\noindent where $y_s$ is the source class and $y_t$ is the target class. In addition to the common label-flipping attacks \cite{DataPoisoning20Tolpegin}, where $y_s\neq y_t$, we also consider label-setting attacks where $y_t=1$ (reporting channels as always busy) or $y_t=0$ (reporting channels as always available).

\begin{table}[b]\smallsize
\caption{Comparison of SSVax with other defenses.}
\label{tab:comparison}\centering
\begin{tabular}{|l|c|c|c|c|c|}\hline
\textbf{Scheme} & \textbf{Majority} & \textbf{FC} & \textbf{Light} & \textbf{Updates} \\
& \textbf{resilience} & \textbf{dataset} & \textbf{SUs} & \textbf{integrity} \\\hline\hline
\Median & \multirow{2}{*}{$\times$} & \multirow{2}{*}{$\times$} & \multirow{2}{*}{\checkmark} & \multirow{2}{*}{$\times$} \\
\TrMean~\cite{TrimmedMean18Yin} & & & & \\\hline
\FLShield~\cite{FLShield24Kabir} & $\times$ & $\times$ & $\times$ & \checkmark \\\hline
MESAS \cite{MESAS23Kraub} & $\times$ & $\times$ & \checkmark & \checkmark \\\hline
\FLTrust~\cite{FLTrust22Cao} & \checkmark & \checkmark & \checkmark & $\times$ \\\hline
Our \SSVax & \checkmark & \checkmark & \checkmark & \checkmark \\\hline
\end{tabular}
\end{table}

Our goal is to design an algorithm that secures FLSS by detecting and filtering malicious updates from compromised SUs. The defense is implemented on the FC side, assuming no prior knowledge of the number of malicious SUs. It relies on a small, clean labeled training dataset ($\mathcal{D}^S$), similar to assumptions made in the literature \cite{FLTrust22Cao} and consistent with practical SS scenarios, where the FC collaborates with PUs for accurate spectrum labels.
While acquiring a trusted dataset at the aggregation server poses challenges in decentralized FL environments, it is more feasible in the SS domain. Spectrum data can be reliably collected within the PUs' coverage area by the FC or SUs, enabling the creation of a clean dataset for robust defense.

Given the assumption that malicious updates deviate significantly from benign ones due to differing objectives \cite{MESAS23Kraub}, we propose \SSVax, a defense mechanism that detects and filters malicious updates by comparing them against pseudo-malicious updates (vaccine) distilled at the FC. Table \ref{tab:comparison} summarizes the key differences between our approach and other SOTA defense mechanisms. Notably, \SSVax\ is independent of the majority assumption, requires only a small dataset at the FC, imposes no additional costs on the SUs, and preserves local updates for high learning performance.

\begin{figure}[t]
    \centering
    \includegraphics[width=1\linewidth]{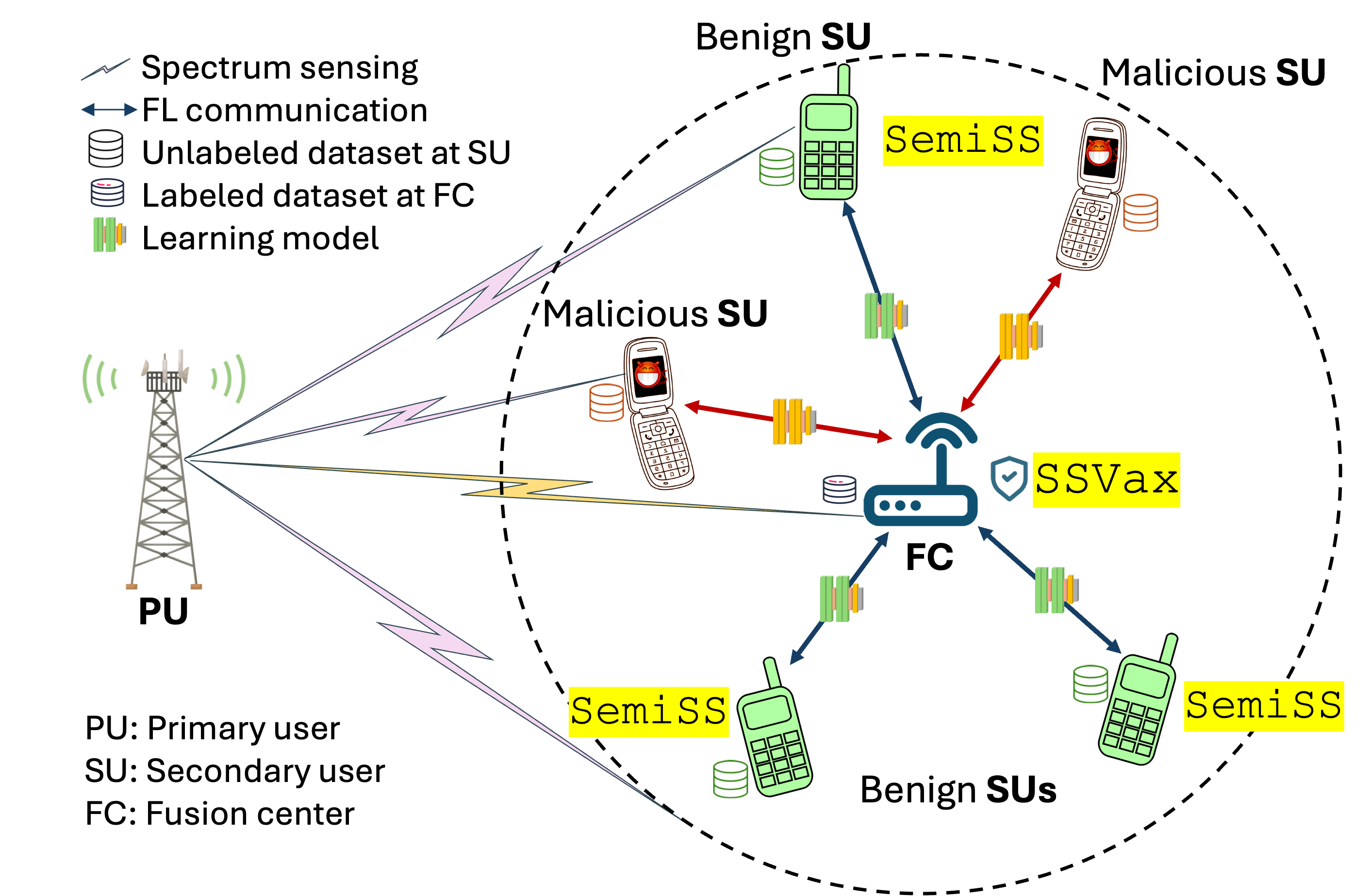}
    \caption{System overview of the FLSS system, where some SUs act maliciously. The proposed \SemiSS\ algorithm runs on SUs to train local models on unlabeled data, while \SSVax\ operates on the FC to secure central aggregation.}
    \label{fig:system-overview}
\end{figure}

\subsection{Scope and limitations}
Fig. \ref{fig:system-overview} presents an overview of the proposed algorithms, \SemiSS\ and \SSVax, which aim to enhance the robustness and security of FLSS systems. Both algorithms operate under the assumption of a small, benign labeled dataset available at the FC. They are particularly tailored for SS in cognitive radio networks. \SemiSS\ employs energy detection for label correction, a technique specific to SS applications. Additionally, the improved performance using frequency-domain data over raw IQ samples is focused on SS applications and may not be directly applicable to other domains. \SSVax\ focuses on mitigating poisoning attacks in FLSS but shows promise for broader applications in securing general FL systems, as explored in the subsequent discussion section.

\section{FL-based Spectrum Sensing for B5G}\label{learning}
Recent advances in SS, such as \DeepSense~\cite{DeepSense21Uvaydov}, have demonstrated the capability to detect PUs at sub-millisecond speeds, enabling real-time, multi-channel detection directly from raw IQ signals. Building on its proven success, \DeepSense\ is adopted as the core ML model in our proposed \SemiSS, specifically designed to tackle the challenge of limited labeled data in FLSS training.

\begin{figure}[t]
    \centering
    \includegraphics[width=.9\linewidth]{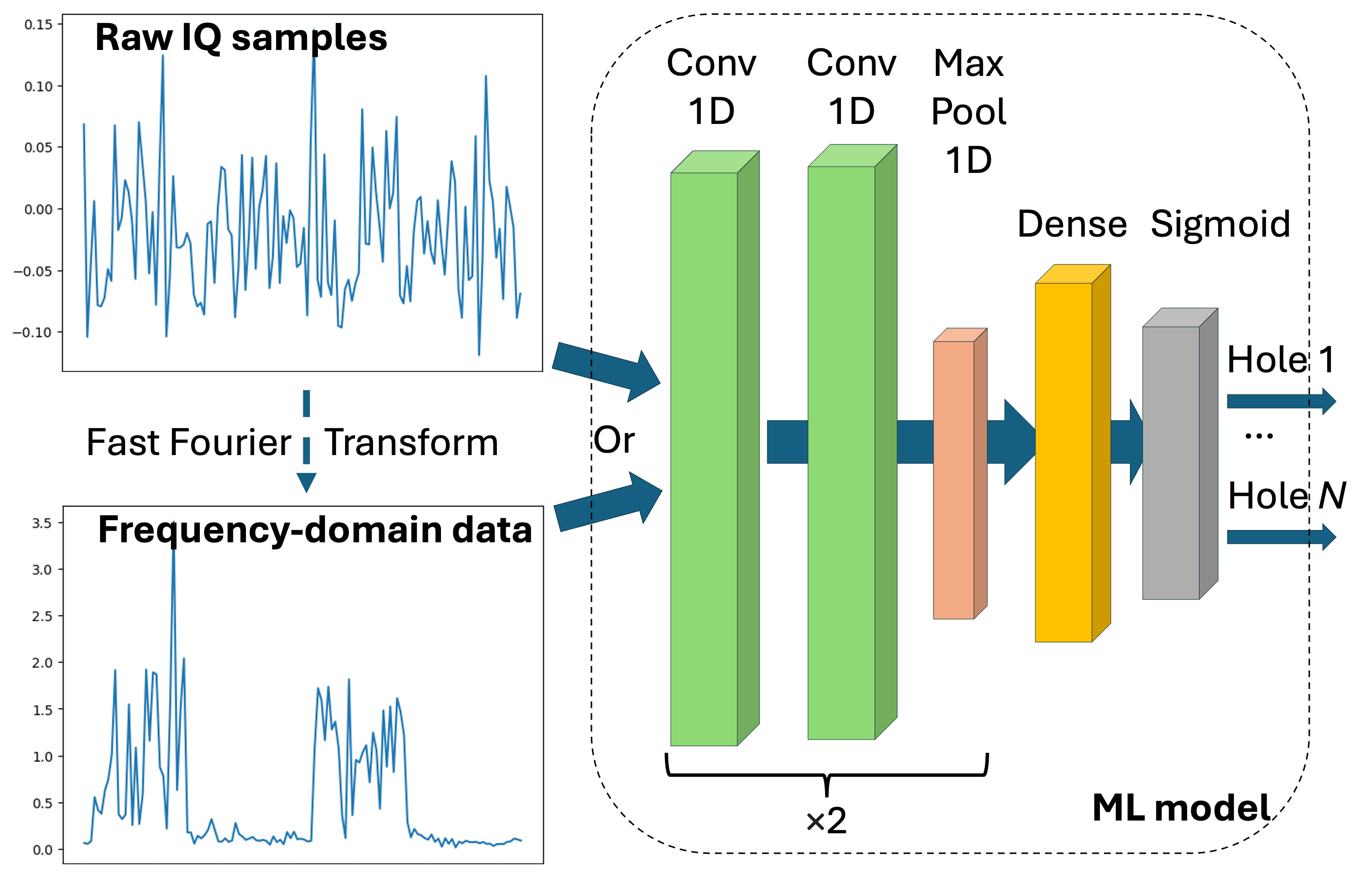}
    \caption{A multi-label CNN architecture for on-device SS.}
    \label{fig:deep-sense}
\end{figure}

\subsection{Preliminary experiments}
\DeepSense\ utilizes a one-dimensional CNN (DNN) to process IQ data received from the SS module (e.g., antennas), outputting multi-label classifications that indicate the PU status across multiple channels. As shown in Fig. \ref{fig:deep-sense}, this architecture processes 32 IQ data points in real-time on FPGA devices, achieving near-perfect detection accuracy \cite{DeepSense21Uvaydov}. Additionally, \DeepSense\ can leverage frequency-domain data obtained by applying an FFT to raw IQ samples \cite{DeepSweep24Robinson}. Our observations suggest that representing radio signals in the frequency domain provides greater visual intuitiveness, aiding tasks such as dividing wideband signals into narrowbands or identifying active PU signals \cite{LabelFree24Milosheski}. Fig. \ref{fig:deep-sense} also displays an example of raw IQ data and its corresponding frequency-domain representation for four channels, labeled as $[1,0,1,0]$, where the first and third channels show active PUs, while the others contain background noise. The frequency-domain representation is more visually intuitive for identifying PU signals, which implies that \emph{ML model performance might improve when using frequency-domain data rather than raw IQ samples}, despite the additional computational overhead of FFT.

\begin{figure}[t]
    \centering
    \begin{subfigure}[b]{.23\textwidth}
        \centering
        \includegraphics[width=\textwidth]{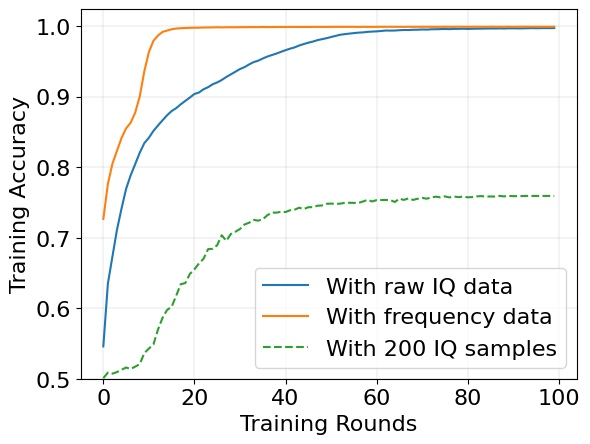}
        \caption{On LTE dataset.}
    \end{subfigure}
    \begin{subfigure}[b]{.23\textwidth}
        \centering
        \includegraphics[width=\textwidth]{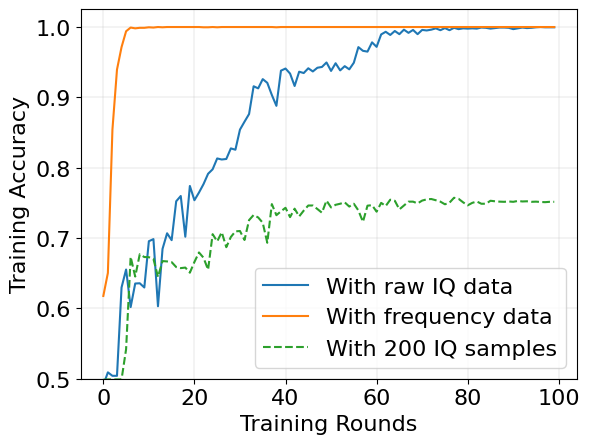}
        \caption{On SDR dataset.}
    \end{subfigure}
    \caption{Learning performance of FLSS across datasets, evaluated with time-domain (IQ) data, frequency-domain data, and a limited sample size (200 samples).}
    \label{fig:fl-learning}
\end{figure}

In our FLSS implementation, \DeepSense\ serves as the learning model, while a central FC coordinates model training across multiple SUs. For preliminary experiments, we simulate $\mathbf{N}=100$ SUs, each receiving a training dataset consisting of sensing data and corresponding PU status labels. Both synthetic and real-world datasets from \cite{DeepSense21Uvaydov} are used: a synthetic LTE dataset with $16$ channels and a real-world SDR dataset with $4$ channels, each containing radio signal indicating PU activity. Further dataset details are provided in the experimental evaluation section. To support dense wireless communication scenarios, such as crowds of mobile devices, we adopt a cross-device FL strategy \cite{FLSurvey23Li}. Each FL round involves a random selection of $\mathbf{R}_S=10\%$ SUs for local training, conducted over $\mathbf{T}=100$ rounds. Results are shown in Fig.~\ref{fig:fl-learning}, with further experimental settings provided in the evaluation section.

Our results successfully replicate the accuracy of \DeepSense\ in a distributed FL setting. Notably, using frequency-domain data instead of raw IQ signals accelerates model convergence. For example, on the LTE dataset, convergence occurs after $17$ rounds, while the SDR dataset reaches $99\%$ accuracy in just $8$ rounds. This faster convergence reduces training overhead, conserving resources on mobile devices.
Additionally, we evaluate the learning performance of FLSS with limited data samples to facilitate our approach to address the critical challenge of acquiring labeled data for local training on SU devices, as presented subsequently.

\subsection{Our \SemiSS}
SOTA ML-based SS approaches are predominantly supervised, relying heavily on extensive labeled datasets. However, methods for automated labeling remain under-explored in the literature. Semi-supervised learning (SSL), widely applied in other domains \cite{DeepClustering24Ren}, presents a promising solution by significantly reducing the need for labeled data. Building on this concept, we propose a semi-supervised FL-based SS approach that incorporates label correction through energy detection, enabling performance comparable to fully labeled FLSS training.

\begin{figure}[t]
    \centering
    \begin{subfigure}[b]{.29\textwidth}
        \centering
        \includegraphics[width=\textwidth]{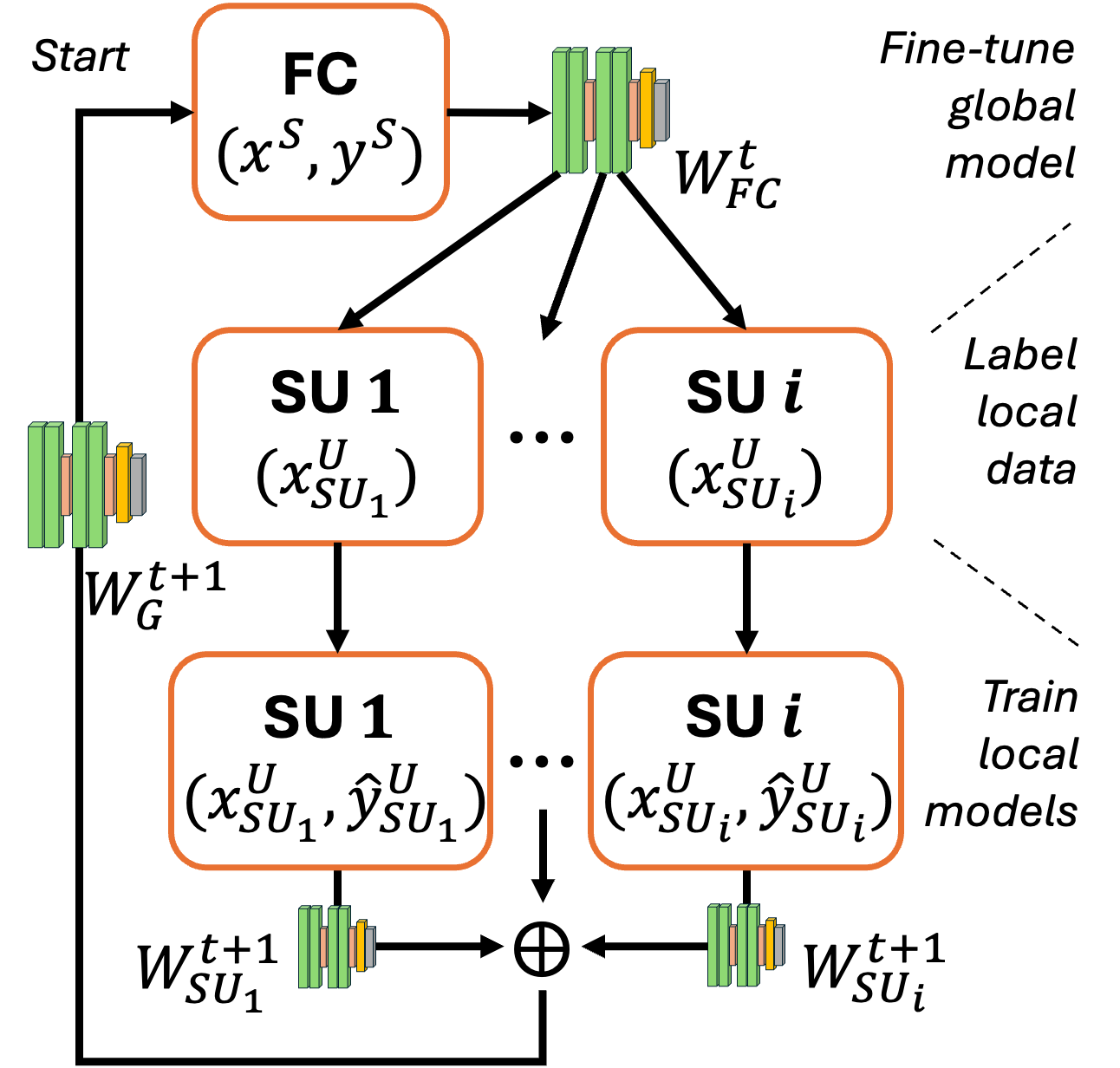}
        \caption{\SemiFL\ based FLSS.}
        \label{fig:semi-FL}
    \end{subfigure}
    \begin{subfigure}[b]{.17\textwidth}
        \centering
        \includegraphics[width=\textwidth]{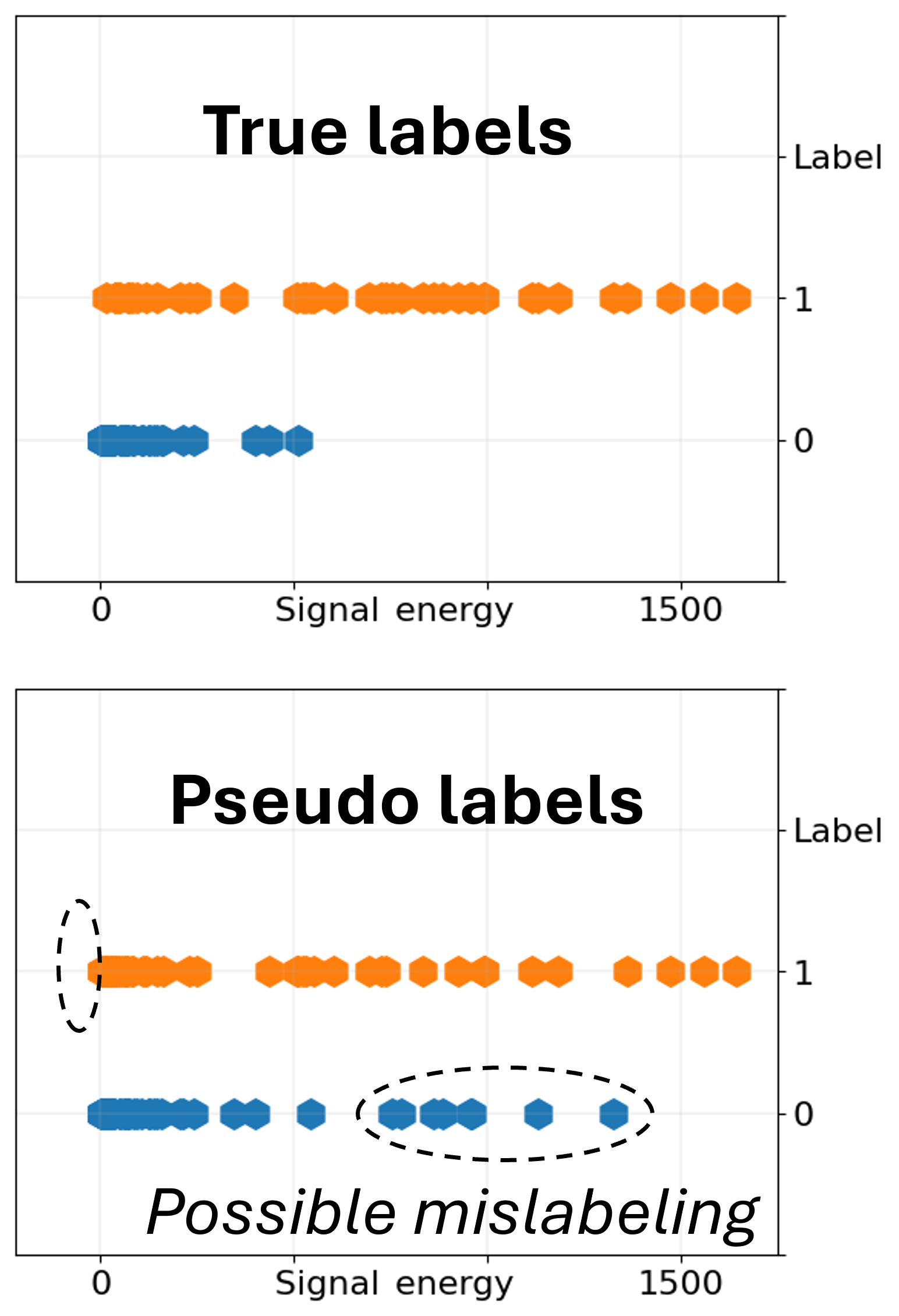}
        \caption{Signal energy vs. \\ labels.}
        \label{fig:energy_detection}
    \end{subfigure}
    \caption{Workflow of semi-supervised FLSS and examples of signal energy.}
\end{figure}

Fig. \ref{fig:semi-FL} illustrates the SOTA semi-supervised FL approach, \SemiFL\ \cite{SemiFL24Diao}. In this method, a small labeled dataset $\mathcal{D}^S$ resides on the server, while the participating clients only hold unlabeled data $\mathcal{D}^U$. The process begins with training a supervised model on $\mathcal{D}^S$, which is then distributed to clients for pseudo-labeling their local data. Clients use these pseudo-labeled datasets for local training before sending the model updates back to the server. The server aggregates the updates and fine-tunes the model on $\mathcal{D}^S$ before the next FL round. While effective in computer vision domains, \SemiFL\ relies heavily on image-based augmentations (e.g., cropping, rotation), which are unsuitable for radio signals as such transformations can degrade SS accuracy.

In our experiments, using $|\mathcal{D}^S|=200$, the initial supervised model $W_{FC}$ achieves an accuracy of $76\%$. When used to pseudo-label the unlabeled data, the labeling accuracy is inherently limited by the performance of $W_{FC}$, resulting in a final accuracy of $78\%$ on the LTE dataset. This highlights the need for improved pseudo-label correction to enhance local learning on unlabeled data.

Energy detection is a conventional method for SS, where the presence of a PU is determined by comparing the received signal energy to a predefined threshold. However, noise and radio propagation effects can result in nonzero energy readings even when no PU is active on a channel, making threshold selection challenging. Fig. \ref{fig:energy_detection} shows energy levels for a channel with and without PU activity, illustrating that no fixed threshold reliably detects PU presence. To address this, we examine pseudo-labeled data based on energy levels and apply corrections to mitigate misclassifications. Specifically:
\begin{itemize}
\item High-energy samples \emph{misclassified} as `no PU' (label 0) are corrected to `PU active' (label $1$).
\item Low-energy samples \emph{misclassified} as `PU active' (label 1) are corrected to `no PU' (label 0).
\end{itemize}
This approach circumvents the rigid thresholding typical of conventional energy detection methods, enabling more accurate pseudo-labeling and improving local training performance.

Motivated by insights gained from preliminary experiments, we propose \SemiSS\ for training FLSS on unlabeled data. The workflow, summarized in Alg. \ref{alg:SSFLSS}, builds on the \SemiFL\ framework by incorporating energy detection to refine pseudo-labels, thereby improving learning performance. The training process proceeds as follows: Initially, the FC trains $W_{FC}$ on its labeled dataset $\mathcal{D}^S$ and distributes the model as $W_G$ to SUs for the first round of FL training. In each subsequent round, upon receiving the global model, the $i$-th SU uses $W_G$ to generate pseudo-labels for its local unlabeled data $\mathcal{D}^U_{SU_i}$. These pseudo-labels, denoted as $\hat{y}$, are refined through energy detection as described by the following rule:

\begin{equation}
\bar{y}=
\begin{cases}
0, & \text{for $C_0$ lowest energy samples labeled $\hat{y}=1$}\\
1, & \text{for $C_1$ highest energy samples labeled $\hat{y}=0$}\\
\hat{y}, & \text{otherwise}
\end{cases},
\label{eq:label_correct}
\end{equation}

\noindent where $C_0$ and $C_1$ represent the number of samples corrected for misclassifications when $\hat{y}=1$ and $\hat{y}=0$, respectively. These values are determined empirically, as discussed in subsequent sections. After refining the pseudo-labels (line $18$), each SU performs local training on its updated dataset and sends the resulting model $W_{SU_i}$ back to the FC. The FC aggregates the models from all selected SUs into a new global model and fine-tunes it using $\mathcal{D}^S$ \cite{SemiFL24Diao} before redistributing the updated model to SUs for the next training round. This iterative process continues until the model achieves the desired accuracy or the predefined number of training rounds completed. In the following section, we analyze the algorithm's computational complexity and discuss optimal parameter settings for label correction.

\begin{algorithm}[t]\smallsize
\caption{The proposed \SemiSS.}\label{alg:SSFLSS}
\SetAlgoLined\SetKwProg{Fn}{}{:}{}
\Fn{System execution} {
    \nl FC initializes the global model $W_G^0$\\
    \nl$W_G^0\gets \textsc{FC\_Update}(W_G^0)$\\
    \nl\For{each training round $t = 1,2,\ldots\mathbf{T}$} {
        \nl$\mathcal{S}^t\gets$ uniformly sample $\mathbf{N}_S$ SUs\\
        \nl\For{each SU$_i\in\mathcal{S}^t$ \textbf{in parallel}} {
            \nl$W_{SU_i}^t\gets \textsc{SU\_Update}(W_G^{t-1})$
        }
        \nl$W_G^{t}\gets 1/\mathbf{N}_S\sum_{SU_i\in\mathcal{S}^t}W_{SU_i}^{t}$\\
        \nl$W_G^t\gets\textsc{FC\_Update}(W_G^t)$
    }
    \nl\Return $W_G^\mathbf{T}$\\
}
\Fn{$\textsc{FC\_Update}(W_G)$} {
    \nl$W_{FC}\gets W_G$\\
    \nl\For{each local epoch from $1$ to $\mathbf{E}$} {
        \nl\For{each data batch $(x,y)\in\mathcal{D}^S$} {
            \nl$L\gets\mathcal{L}(f(x,W_{FC}),y)$\\
            \nl$W_{FC}\gets W_{FC}-\eta\nabla_WL$
        }    
    }
    \nl\Return $W_{FC}$
}
\Fn{$\textsc{SU\_Update}(W_G)$} {
    \nl$W_{SU}\gets W_G$\\
    \nl$\hat{y}\gets f(x,W_{G})$\\
    \nl$y\gets$ correct the pseudo-labels $\hat{y}$\\
    \nl\For{each local epoch from $1$ to $\mathbf{E}$} {
        \nl Apply lines 10--12 from Algorithm \ref{alg:FLbasic}
    }
    \nl\Return $W_{SU}$ to FC
}
\end{algorithm}

\subsection{Complexity analysis}
The proposed \SemiSS\ introduces a spectrum energy calculation for each label, a straightforward process that can be efficiently performed as a preprocessing step. Identifying the least and most energetic samples involves sorting, which has a time complexity of $O(n^2)$. As a result, the overall runtime complexity of \SemiSS\ remains comparable to that of \SemiFL.

In signal theory, PU signals typically exhibit higher energy levels than background noise, which allows for the use of a threshold to distinguish PU activity (label $1$) from inactivity (label $0$). Unlike traditional methods, our approach refines pseudo-labels by targeting the least and most energetic samples for correction, eliminating the need for a fixed threshold. Instead, we employ a dynamic strategy--\emph{how many samples should be corrected?}--which acts as a `soft threshold' that adjusts throughout the training process.

As training progresses and model accuracy improves, the number of corrected samples can be gradually reduced. To implement this, we tie the soft threshold to the supervised FC model's error rate. This dynamic adjustment ensures a balance:
\begin{itemize}
\item If the threshold exceeds the FC model's error rate, it may lead to the unnecessary correction of correctly labeled samples, negatively impacting learning performance.
\item Conversely, if too few samples are corrected, the model may not fully benefit from pseudo-label refinement, limiting improvements in accuracy.
\end{itemize}
Our experimental evaluation (discussed later) demonstrates that this dynamic adjustment allows \SemiSS\ to achieve performance comparable to training with a fully labeled dataset.

\section{Secured FL-based Spectrum Sensing}\label{security}
FLSS leverages FL for model training, which inherently inherits FL's vulnerabilities to security threats. To strengthen the security of FLSS, we investigate the impact of data poisoning attacks and the limitations of existing defense mechanisms. Building on these insights, we propose a novel defense strategy, inspired by vaccination, designed to enhance the resilience of FLSS against malicious attacks.

\begin{figure}[t]
    \centering
    \begin{subfigure}[b]{.23\textwidth}
        \centering
        \includegraphics[width=\textwidth]{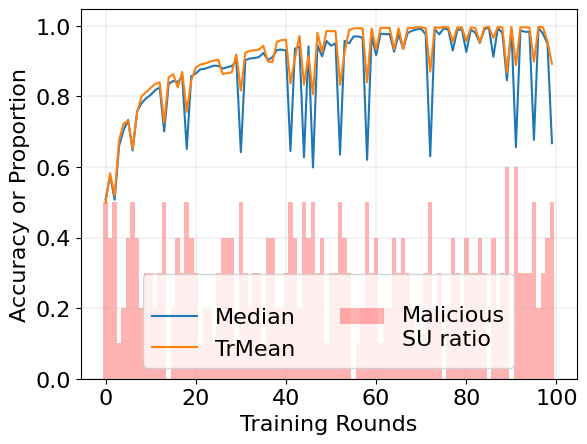}
        \caption{With \Median\ and \TrMean}
        \label{fig:median-tr_mean}
    \end{subfigure}
    \begin{subfigure}[b]{.23\textwidth}
        \centering
        \includegraphics[width=\textwidth]{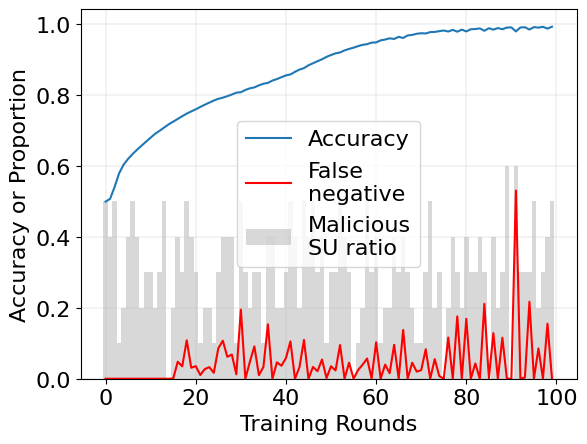}
        \caption{With \FLTrust.}
        \label{fig:with_fltrust}
    \end{subfigure}
    \caption{Learning accuracy of FLSS with SOTA defenses under label-flipping attacks.}
    \label{fig:robust_byzantine}
\end{figure}

\subsection{Attacks and defenses}
Based on the threat model depicted in Fig. \ref{fig:system-overview}, we conduct experiments to evaluate FLSS's susceptibility to data poisoning attacks and the effectiveness of existing defenses. In our setup, malicious SUs execute label-flipping attacks aimed at degrading the accuracy of the global model, while defenses are implemented at the FC.

In our experiments, $\mathbf{R}_M=30\%$ of the SUs are designated as malicious, and $\mathbf{R}_S=10\%$ of the total SUs are randomly selected for each FL training round. Fig. \ref{fig:robust_byzantine} illustrates the accuracy of FLSS with SOTA defenses such as \Median, \TrMean, and \FLTrust. Despite the malicious participant ratio being $30\%$, the random selection of SUs in cross-device FL sometimes results in more than $30\%$ of malicious participants in a given round, with occasional spikes exceeding $50\%$. This randomness exposes a significant limitation of many SOTA defenses, such as MESAS and \FLShield, which assume that benign participants always constitute the majority (over $50\%$).

In a poisoning attack, the FL server receives a combination of benign and malicious updates. Robust aggregation defenses typically assume that benign updates are similar, while malicious updates exhibit significant deviations due to conflicting objectives. Updates are clustered based on their similarities, and the smaller group, presumed to contain malicious updates, is excluded from aggregation \cite{MESAS23Kraub}. However, as shown in Fig. \ref{fig:median-tr_mean}, when malicious updates exceed $50\%$, these defenses fail, resulting in a substantial degradation of model accuracy.

\FLTrust, a recent SOTA method that does not rely on the majority assumption, shows greater resilience in our experiments. It assigns a trust score to each update based on its similarity to the server's update, which determines the weight of contributions during aggregation. As illustrated in Fig. \ref{fig:with_fltrust}, \FLTrust\ maintains consistent performance even with a high ratio of malicious updates. However, deeper analysis reveals that malicious updates can still affect the model, particularly in later training rounds, due to their non-negligible weights in the aggregation process--referred to here as the false negative rate.
The non-zero false negative rate highlights the limitation of using a single benign update to classify client updates, underscoring the need for a more robust mechanism to separate benign from malicious updates.
These findings suggest that while current defenses are effective to some extent, they remain vulnerable when the majority assumption is violated. In response, we propose a novel approach to secure FLSS in the following.

\subsection{Our \SSVax}
To address the vulnerabilities of defenses that rely on the assumption of a benign majority, we propose \SSVax, a novel defense mechanism that eliminates the need for such an assumption. Inspired by the concept of vaccination, \SSVax\ strengthens the robustness of FL by proactively filtering out malicious updates.

In the context of vaccination, the human body is protected from viral infections (e.g., Pfizer or Moderna vaccines for COVID-19) by stimulating the immune system to recognize and neutralize threats. Similarly, \SSVax\ introduces a `vaccinated aggregator' which functions as the immune system of the FL process, detecting and filtering harmful updates to safeguard the global model. The core idea behind \SSVax\ is to generate a distilled `vaccine' on the FC, derived from simulated attacks. This vaccine, referred to as pseudo-malicious updates, is trained using the FC's labeled dataset $\mathcal{D}^S$, which is readily available in our \SemiSS\ setting. Leveraging these pseudo-malicious updates, \SSVax\ classifies incoming updates as either benign or malicious before aggregation. The FC clusters updates based on their similarity to either benign or malicious patterns. Benign updates are aggregated to enhance the global model, while malicious updates are isolated and analyzed to gain insights into attacker behavior. This dual functionality ensures the security of FLSS without relying on a majority of benign participants. The following section outlines the detailed algorithm for \SSVax, followed by an analysis of its performance and implications.


\subsubsection{The proposed algorithm}
Building on the vaccination concept, \SSVax\ strengthens the FLSS system by protecting against malicious SUs. Using a small, trusted dataset, the FC generates `vaccines' to detect and filter out malicious updates. These vaccines act as anchors for classification, establishing a clearer boundary between benign and malicious updates. The training procedure for \SSVax\ is detailed in Alg. \ref{alg:FLVax}.

\begin{algorithm}[t]\smallsize
\caption{The proposed \SSVax.}\label{alg:FLVax}
\SetAlgoLined\SetKwProg{Fn}{}{:}{}
\Fn{System execution} {
    \nl Apply lines 1--2 from Algorithm \ref{alg:SSFLSS}\\
    \nl\For{each training round $t = 1,2,\ldots\mathbf{T}$} {
        \nl$\mathcal{S}^t\gets$ uniformly sample $\mathbf{N}_S$ SUs\\
        \nl\For{each SU$_i\in \mathcal{S}^t$ \textbf{in parallel}} {
            \nl$W_{SU_i}^t\gets \textsc{SU\_Update}(W_G^{t-1})$
        }
        \nl\For{each vaccine $Vax_j$ \textbf{in parallel}} {
            \nl$W_{Vax_j}^t\gets$ pseudo-malicious updates
        }
        \nl$\mathcal{W}^t\gets\textsc{FC\_Filter}\left(\{W_{SU_i}^t\},\{W_{Vax_j}^t\}\right)$\\
        \nl$W_G^{t}\gets |\mathcal{W}^t|^{-1}\sum_{W_i\in\mathcal{W}^t}W_i$\\
        \nl Apply line 8 from Algorithm \ref{alg:SSFLSS}
    }
    \nl\Return $W_G^\mathbf{T}$\\
}
\Fn{$\textsc{FC\_Filter}(\mathcal{W}_{SU},\mathcal{W}_{Vax})$} {
    \nl$\mathcal{W}\gets\mathcal{W}_{SU}\cup\mathcal{W}_{Vax}$\\
    \nl Cluster $\mathcal{W}$ into $|\mathcal{W}_{Vax}|+1$ groups, with members of $\mathcal{W}_{Vax}$ fixed as cluster centroids\\
    \nl Return the cluster not containing any member of $\mathcal{W}_{Vax}$
}
\end{algorithm}

The procedures at local SUs remain consistent with those in Alg.~\ref{alg:SSFLSS} for FLSS; however, the server-side operations are enhanced with the vaccine mechanism. This includes \emph{vaccine distillation} (lines $6-7$) and \emph{vaccine injection} into the system (line $8$). The vaccine, represented by pseudo-malicious updates, is employed to detect incoming malicious updates through the $\textsc{FC\_Filter}$ function. This detection step occurs prior to aggregation, ensuring that only benign updates contribute to the global model. \SSVax\ provides a flexible framework, allowing for customization of both the vaccine design and the detection mechanism.

A pseudo-malicious update (vaccine) is generated by training the learning model with a pseudo-attack designed to mimic the targeted attack.
Selecting an appropriate vaccine is crucial for effective defense, as each vaccine is tailored to counter specific threat types. \SSVax\ leverages the common observation that benign updates are typically similar and converge during training, whereas malicious updates deviate due to conflicting objectives. Therefore, the defense's effectiveness depends on how well the pseudo-malicious updates correspond to the actual type of attack. For example, in scenarios involving targeted attacks where real and pseudo-attacks target different channels, the defense's performance may vary accordingly.

The primary goal of clustering in \SSVax\ is to classify incoming updates as either benign or malicious. Given $\mathbf{N}_P$ pseudo-malicious updates and $\mathbf{N}_S$ incoming updates, all pseudo-malicious updates are fixed as cluster centroids. Ideally, the clustering process results in $\mathbf{N}_P+1$ groups, where incoming malicious updates are grouped with the pseudo-malicious centroids, while benign updates form a separate group. The benign updates are then forwarded for aggregation into the global model. \SSVax\ employs a customized \Kmeans\ algorithm for clustering due to its simplicity and effectiveness. However, exploring alternative clustering methods represents a promising avenue for future research. Alg. \ref{alg:Kmeans} presents the tailored \Kmeans\ algorithm, which clusters $n$ data points into $k+1$ clusters, where $k$ centroids are pre-defined. We next analyze the complexity of \SSVax\ and its compatibility with other defense mechanisms.

\begin{algorithm}[t]\smallsize
\caption{Customized \Kmeans\ clustering.}\label{alg:Kmeans}
\SetAlgoLined\SetKwProg{Fn}{}{:}{}
\SetKwFor{Repeat}{repeat}{}{until~\normalfont{convergence criteria are met}}
\Fn{Input} {
    $\mathcal{W}_p$: set of $n$ data points\\
    $\mathcal{W}_c$: set of $k$ fixed centroids
}
\Fn{Output} {
    A set of $k+1$ clusters
}
\Fn{Function execution} {
    \nl Choose $\mathcal{W}_c$ and the mean of $\mathcal{W}_p$ as initial centroids\\
    \Repeat{}{
        \nl Assign each point in $\mathcal{W}_p$ to the cluster with the closest centroid\\
        \nl Update the centroid of the $(k+1)^{\text{th}}$ cluster based on its assigned points\\
    }
    \nl\Return $k+1$ clusters
}
\end{algorithm}

\subsubsection{Algorithm analysis}
The SU-side training process in \SSVax\ remains consistent with standard FL protocols, while the FC incorporates additional steps for vaccine distillation and malicious update filtering. In typical FL scenarios, servers are expected to have sufficient computational resources to handle participant coordination, communication, and aggregation. The computational overhead of generating pseudo-malicious updates on the FC is therefore manageable, especially when compared to the resource constraints faced by SU devices. Additionally, the training of pseudo-malicious models can occur independently and in parallel with SU training and other FC operations, thereby minimizing latency.

\SSVax\ requires a small, trusted labeled dataset at the FC, similar to \FLTrust, whereas defenses like \FLShield\ and MESAS do not. However, a significant advantage of \SSVax\ is its independence from the majority-benign assumption relied upon by many SOTA defenses. By leveraging clustering results, \SSVax\ can estimate the proportion of malicious participants, offering valuable insights for monitoring and managing the learning federation, thus advancing explainable AI \cite{ExplainableAI23Ali}. Table \ref{tab:comparison} provides a comparison between \SSVax\ and current SOTA methods, emphasizing that \SSVax\ does not rely on majority assumptions, requires only a minimal dataset at the FC, imposes no additional burden on SUs, preserves benign updates to maintain FL learning performance, and enhances explainability.


\SSVax\ is highly flexible, allowing customization of vaccine distillation methods and clustering algorithms to address specific attack scenarios. Moreover, it is compatible with other defenses mechanisms for enhanced security. For example, filtered updates from \SSVax\ can be further processed using \FLTrust\ to mitigate the influence of stealthy malicious updates. Metrics from MESAS could enhance the clustering algorithm's ability to differentiate between benign and malicious updates. Similarly, having a small benign dataset on the server enables additional validation using per-class loss metrics from \FLShield\ to counter backdoor attacks. Based on this analysis, the following section presents comprehensive experiments to evaluate \SSVax, compare its performance, and demonstrate its integration with other techniques to effectively secure FLSS systems.

\section{Experiment Evaluation}\label{evaluation}
This section evaluates the proposed FLSS framework in terms of learning performance and its resilience against both untargeted and targeted data poisoning attacks.

\subsection{Experiment setup}
The experiments leverage datasets provided by \cite{DeepSense21Uvaydov} for SS applications. These include a synthesized LTE dataset and a real-world SDR dataset: The LTE dataset contains transmissions across 16 sub-bands (channels) and is generated using MATLAB's LTE Toolbox to simulate LTE-M uplink transmissions over a $10$-MHz-wide band. To increase classification complexity, additive white Gaussian noise and Rayleigh fading are introduced, simulating realistic environment conditions. The real-world SDR dataset is collected using five software-defined radios (SDRs) operating with GNU Radio. Four SDRs function as transmitters and one serves as a receiver. Data was collected over two days with varying transmitter orientations to incorporate diverse channel effects. The dataset covers transmissions across four non-overlapping channels, each $5$ MHz wide, resulting in a total bandwidth of $20$ MHz. Further technical details about these datasets can be found in~\cite{DeepSense21Uvaydov}.

For model training, the \DeepSense's DNN model is used, configured for FL under a cross-device strategy. The general experimental setup includes: A total of $100$ simulated UEs uniformly assigned to the unlabeled dataset $\mathcal{D}^U$. The FC operates with a small labeled dataset $\mathcal{D}^S$ containing $200$ samples. Each FL training round proceeds as follows:
\begin{itemize}
\item The FC selects $\mathbf{R}_S=10\%$ of the total UEs for training participation.
\item $\mathbf{R}_M(\%)$ of the UEs are predesignated as malicious and execute poisoning attacks when selected for training.
\end{itemize}
\noindent The remaining hyper-parameters are configured as follows: learning rate is set to $1e-3$, number of FL training rounds is $100$, and each UE performs $2$ epochs of local training per round. The FC performs $100$ epochs of supervised training, fine-tunes the model for $2$ epochs, and trains pseudo-malicious models for $1$ epoch during each FL round.

\subsection{Learning performance}
\begin{figure}[t]
    \centering
    \begin{subfigure}[b]{.23\textwidth}
        \centering
        \includegraphics[width=\textwidth]{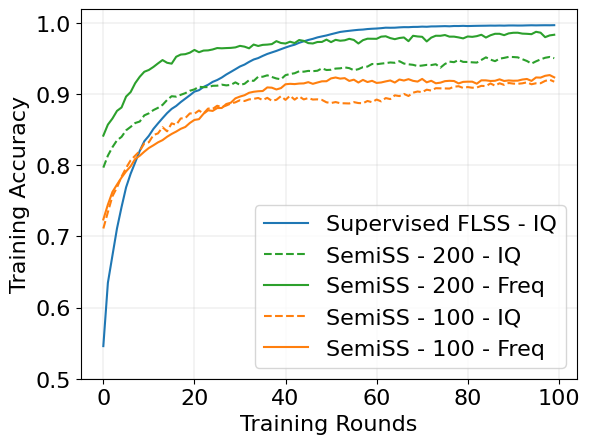}
        \caption{LTE dataset.}
        \label{fig:semi_lte_accuracy}
    \end{subfigure}
    \begin{subfigure}[b]{.23\textwidth}
        \centering
        \includegraphics[width=\textwidth]{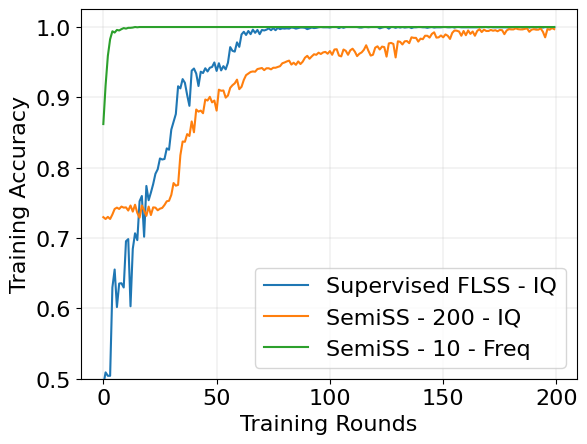}
        \caption{SDR dataset.}
        \label{fig:semi_sdr_accuracy}
    \end{subfigure}
    \caption{Learning performance of \SemiSS\ on two datasets.}
    \label{fig:semi_accuracy}
\end{figure}

We evaluate the learning accuracy of \SemiSS\ on both the LTE and SDR datasets under various hyper-parameter settings and dataset distributions. Fig. \ref{fig:semi_accuracy} presents the accuracy of \SemiSS\ across training rounds compared to a fully supervised FLSS. Overall, \SemiSS\ demonstrates higher initial accuracy due to the supervised pre-training of the FC model. By leveraging label correction during local training, \SemiSS\ achieves final accuracy comparable to that of a fully supervised FLSS trained on completely labeled datasets. Remarkably, \SemiSS\ requires only $200$ labeled samples for both datasets, corresponding to just $0.04\%$ of the total LTE dataset size and $0.3\%$ of the total SDR dataset size.

The results in Fig. \ref{fig:semi_accuracy} also compare the performance of \SemiSS\ using IQ and frequency domain data. For the LTE dataset, frequency domain processing consistently yields higher learning accuracy. Similarly, for the SDR dataset, it significantly accelerates convergence. These findings highlight the advantages of processing frequency domain data over raw IQ samples in SS applications. Next, we investigate the impact of varying hyper-parameter settings on the learning performance of \SemiSS.

\begin{table}[b]\centering\smallsize
\caption{Learning performance of \SemiSS\ on the LTE dataset under varying hyper-parameter settings.}\label{tab:flss_lte_acc}
\begin{tikzpicture}
\node (table) at (0,0) {
\begin{tabular}{|c|c|c|c|c|c|c|}\cline{6-7}
\multicolumn{5}{c|}{} & \multicolumn{2}{c|}{\textbf{Accuracy on}} \\\hline
$|\mathcal{D}^S|$ & $\mathbf{N}$ & $\mathbf{R}_S$ & $\mathbf{E}$ & $\mathbf{R}_C$ & \textbf{Freq data} & \ \ \textbf{IQ data} \ \ \\\hline\hline
$300$ & \multirow{3}{1.3em}{$100$} & \multirow{3}{1em}{$0.1$} & \multirow{3}{1em}{~$2$} & \multirow{3}{1.3em}{$0.3$} & $\mathbf{98.7\%}$ & $\mathbf{95.5\%}$ \\\cline{1-1}\cline{6-7}
$200$ & & & & & $98.4\%$ & $95.1\%$ \\\cline{1-1}\cline{6-7}
$100$ & & & & & $92.3\%$ & $91.9\%$ \\\hline\hline
\multirow{2}{1.3em}{$200$} & $200$ & \multirow{2}{1em}{$0.1$} & \multirow{2}{1em}{~$2$} & \multirow{2}{1.3em}{$0.3$} & $\mathbf{96.8\%}$ & $\mathbf{91.6\%}$ \\\cline{2-2}\cline{6-7}
& $300$ & & & & $92.9\%$ & $89.9\%$ \\\hline\hline
\multirow{2}{1.3em}{$200$} & \multirow{2}{1.3em}{$100$} & $0.2$ & \multirow{2}{1em}{~$2$} & \multirow{2}{1.3em}{$0.3$} & $98.9\%$ & $95.0\%$ \\\cline{3-3}\cline{6-7}
& & $0.3$ & & & $\mathbf{99.1\%}$ & $\mathbf{95.7\%}$ \\\hline\hline
\multirow{2}{1.3em}{$200$} & \multirow{2}{1.3em}{$100$} & \multirow{2}{1em}{$0.1$} & $1$ & \multirow{2}{1.3em}{$0.3$} & $96.8\%$ & $92.6\%$ \\\cline{4-4}\cline{6-7}
& & & $3$ & & $\mathbf{98.1\%}$ & $\mathbf{95.1\%}$ \\\hline\hline
\multirow{3}{1.3em}{$200$} & \multirow{3}{1.3em}{$100$} & \multirow{3}{1em}{$0.1$} & \multirow{3}{1em}{~$2$} & $0.1$ & $98.5\%$ & $94.3\%$ \\\cline{5-7}
& & & & $0.5$ & $\mathbf{98.6\%}$ & $\mathbf{94.9\%}$ \\\cline{5-7}
& & & & $1.0$ & $95.9\%$ & $88.7\%$ \\\hline
\end{tabular}};
\end{tikzpicture}
\end{table}

Table \ref{tab:flss_lte_acc} summarizes the learning performance of \SemiSS\ on the LTE dataset under varying hyper-parameter configurations. The parameters include the size of the labeled dataset, $|\mathcal{D}^S|$, the total number of SUs, $\mathbf{N}$, the SU selection rate, $\mathbf{R}_S$, the number of local training epochs, $\mathbf{E}$, and the label correction ratio, $\mathbf{R}_C$. Both IQ and frequency domain data are evaluated, revealing key insights:

\begin{itemize}
\item A larger labeled dataset at the FC markedly boosts initial accuracy and supports more effective fine-tuning of the global model, leading to improved final accuracy.
\item A smaller local unlabeled dataset (with more participants) decreases learning accuracy, as local training becomes more prone to overfitting individual local data.
\item Increasing the participant selection rate extends the duration of each training round by involving more SUs. While this slightly improves accuracy, it comes at the expense of increased training overhead.
\item The label correction ratio, a pivotal hyper-parameter in the proposed energy detection mechanism, directly influences learning performance. A correction ratio of $100\%$ of the FC model's error rate notably diminishes final accuracy due to the introduction of errors from overly aggressive corrections. Conversely, smaller correction ratios lead to slower convergence and additional training overhead. Experimental results suggest that correction ratios of $0.3$ or $0.5$ achieve an optimal balance, delivering robust performance within $100$ FL rounds.
\end{itemize}

Further experiments varying the FC's labeled dataset distributions (detailed in Appendix \ref{sec:data_distributions}) confirm that the proposed approach consistently achieves perfect detection accuracy when processing frequency-domain data.
In summary, \SemiSS\ demonstrates its capability to support FLSS learning in realistic scenarios involving unlabeled local data, achieving performance comparable to fully supervised learning, while requiring only a small amount of labeled data at the federated coordinator (FC). Next, we evaluate the proposed defense mechanism designed to enhance the security of FLSS systems.

\begin{figure}[t]
    \centering
    \begin{subfigure}[b]{.23\textwidth}
        \centering
        \includegraphics[width=\textwidth]{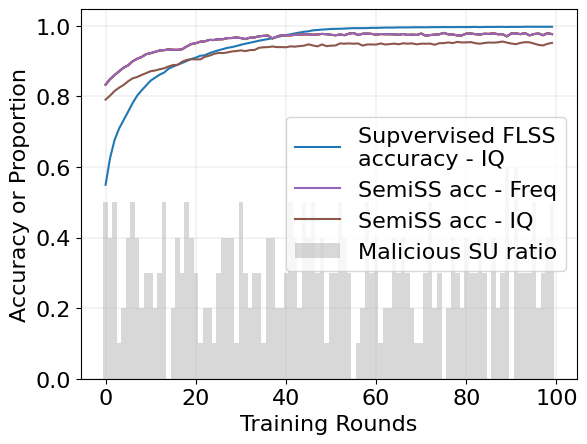}
        \caption{$30\%$ label-flipping attacks.}
        \label{fig:lte_vax_untargeted}
    \end{subfigure}
    \begin{subfigure}[b]{.23\textwidth}
        \centering
        \includegraphics[width=\textwidth]{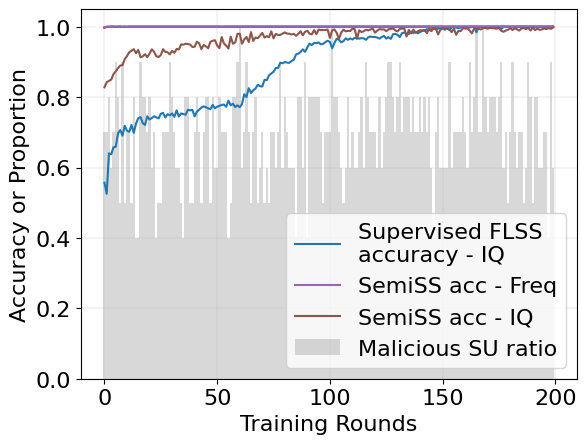}
        \caption{$70\%$ label-setting attacks.}
        \label{fig:sdr_vax_untargeted}
    \end{subfigure}
    \caption{\SSVax\ performance against untargeted attacks: (a) Label-flipping vaccine on LTE dataset and (b) Label-setting vaccine on SDR dataset. 
    }
    \label{fig:LTE_VaX_flipping}
\end{figure}

\subsection{Security evaluation}
This section assesses the effectiveness of \SSVax\ in defending against both untargeted and targeted data poisoning attacks, including comparisons with related approaches and considerations for potential integration with complementary techniques.

\subsubsection{Preventing untargeted attacks}
The first evaluation examines \SSVax's ability to mitigate untargeted label-flipping attacks in FLSS on LTE dataset. To achieve this, the label-flipping vaccine is distilled using pseudo label-flipping on a server dataset containing $200$ samples. The results, shown in Fig. \ref{fig:lte_vax_untargeted}, highlight key metrics across FL rounds, including the ratio of malicious updates, false negative and false positive detection rates, and learning accuracy. \SSVax\ achieves $100\%$ true positive detection with zero false negatives, effectively identifying and removing all malicious updates without mistakenly filtering out benign ones. This preserves learning performance, ensuring convergence comparable to attack-free scenarios. A parallel experiment on the SDR dataset, shown in Fig. \ref{fig:sdr_vax_untargeted}, evaluates the efficacy of the label-setting vaccine in mitigating untargeted label-setting attacks. Even with a high malicious SU ratio of $70\%$, \SSVax\ maintains $100\%$ detection of malicious updates with zero false negatives, safeguarding benign updates and learning accuracy. Across both datasets, \SSVax\ consistently prevents untargeted attacks without compromising the performance of benign updates or learning accuracy. Its robustness is evident across both supervised FLSS and \SemiSS\ settings, achieving performance comparable to attack-free baselines. For clarity and generalizability, results for supervised FLSS on IQ data are presented, as they visually capture the overall trends.

We replicate the experiment in Fig. \ref{fig:lte_vax_untargeted}, increasing the number of SUs from $100$ to $500$ to simulate extremely dense environments, like lecture halls or stadiums. \SSVax\ continues to exhibit the same robust protection. Furthermore, Appendix~\ref{sec:multiple_vaccines} presents \SSVax's use of multiple vaccines to defend against diverse untargeted label-poisoning attacks, demonstrating its high efficacy across various settings.

\begin{figure}[t]
    \centering
    \begin{subfigure}[b]{.23\textwidth}
        \centering
        \includegraphics[width=\textwidth]{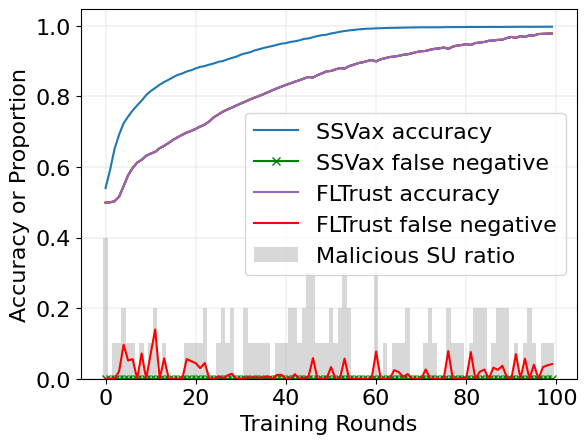}
        \caption{$10\%$ label-setting attacks.}
        \label{fig:lte_vax_targeted}
    \end{subfigure}
    \begin{subfigure}[b]{.23\textwidth}
        \centering
        \includegraphics[width=\textwidth]{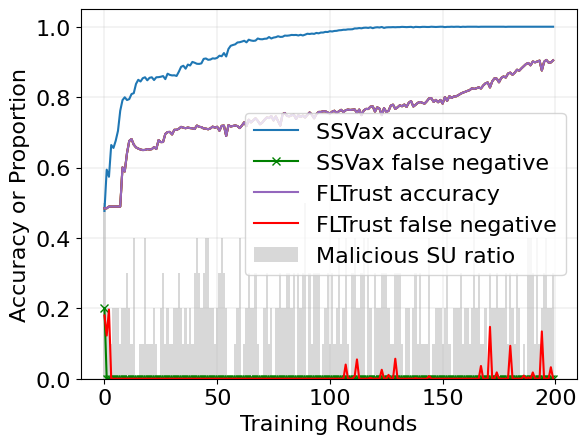}
        \caption{$20\%$ label-flipping attacks.}
        \label{fig:sdr_vax_targeted}
    \end{subfigure}
    \caption{\SSVax\ performance against targeted attacks on half the channels: (a) Label-setting vaccine on LTE dataset, and (b) Label-flipping vaccine on SDR dataset.}
    \label{fig:VaX_targeted}
\end{figure}

\subsubsection{Preventing targeted attacks}
Targeted attacks are stealthier than untargeted ones, as they typically cause minimal degradation in learning performance and accuracy, making the impact less noticeable. These attacks focus on specific channels, reducing the disparity between malicious and benign updates, thereby increasing their stealthiness. In this section, we evaluate \SSVax\ against targeted attacks on both datasets, configuring the attack to target half of the channels. The results are presented in Fig.~\ref{fig:VaX_targeted}. Key findings indicate that \SSVax\ successfully identifies and removes the majority of malicious updates, providing robust defense against targeted attacks. In the absence of malicious updates, \SSVax\ achieves zero false negatives, thereby maintaining optimal learning performance. During the early rounds, some malicious updates evade detection due to the divergence of benign updates and the inherently stealthy nature of targeted attacks. This limitation will be explored in more detail subsequently.

To provide a comparative perspective, we also include the performance of \FLTrust\ in defending against targeted attacks. 
\FLTrust, which scales updates during aggregation, exhibits slower convergence and lower learning accuracy compared to \SSVax. For instance, in the LTE experiment, \SSVax\ reaches maximum accuracy within $60$ rounds, while \FLTrust\ requires over $100$ rounds. The difference in convergence rate is even more pronounced with the SDR dataset. Additionally, \FLTrust\ allows some malicious updates to persist during aggregation (as indicated by the red lines in Fig. \ref{fig:VaX_targeted}), whereas \SSVax\ effectively filters out most malicious updates, achieving faster and more stable convergence. Next, we analyze the impact of varying the number of targeted channels, further highlighting \SSVax's strengths and identifying its limitations.

\subsubsection{Ablation study}
In this section, we evaluate \SSVax's detection performance by varying the number of targeted channels in label-flipping attacks. Experiments on the LTE dataset are conducted by targeting between half of the channels ($8$ out of $16$) and a single channel. The primary focus is on the proportion of undetected malicious updates, particularly in the later stages of training. The results, summarized in Table \ref{tab:FLVax_vary_channels}, show that \SSVax\ maintains near-perfect detection when up to $6$ out of $16$ channels are targeted. However, as the number of targeted channels decreases, the attacks become stealthier, leading to an increase in false negative rates. At higher attack ratios (e.g., $30\%$), \SSVax\ struggles to defend against highly stealthy attacks, especially when only $1$ or $2$ channels are targeted. Nevertheless, in more realistic scenarios with lower attack ratios, \SSVax\ performs robustly, successfully filtering out up to half of the malicious updates.

\begin{table}[t]\centering\smallsize
\caption{False negative rates of \SSVax\ under varying numbers of targeted channels.}\label{tab:FLVax_vary_channels}
\begin{tabular}{|c|c|c|c|}\cline{2-4}
\multicolumn{1}{c|}{} & \multicolumn{3}{c|}{\textbf{Attack ratio}} \\\hline
\textbf{Targeted channels} & $50\%$ & $30\%$ & $10\%$ \\\hline\hline
$8/16$ ($50\%$) & $0\%$ & $0\%$ & $0\%$ \\\hline
$6/16$ ($38\%$) & $3\%$ & $0\%$ & $7\%$ \\\hline
$4/16$ ($25\%$) & $2\%$ & $14\%$ & $24\%$ \\\hline
$2/16$ ($13\%$) & $89\%$ & $77\%$ & $45\%$ \\\hline
$1/16$ ($06\%$) & $99\%$ & $99\%$ & $43\%$ \\\hline
\end{tabular}
\end{table}

\begin{figure}[t]
    \centering
    \begin{subfigure}[b]{.23\textwidth}
        \centering
        \includegraphics[width=\textwidth]{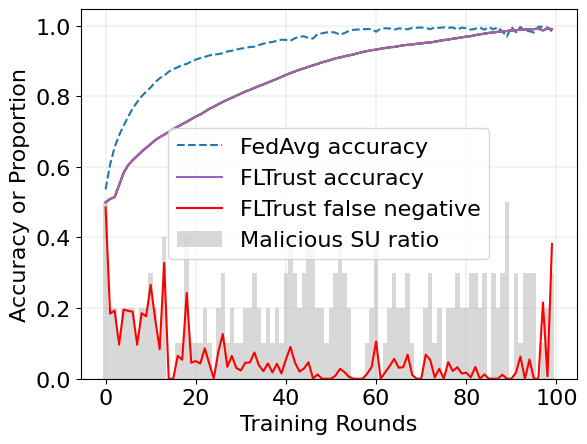}
        \caption{\FLTrust}\label{fig:targeted_one_trust}
    \end{subfigure}
    \begin{subfigure}[b]{.23\textwidth}
        \centering
        \includegraphics[width=\textwidth]{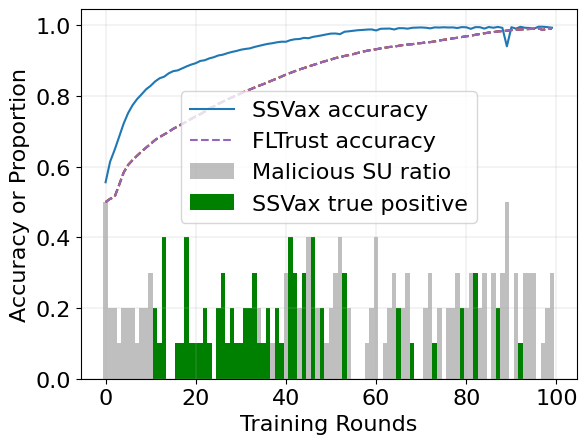}
        \caption{Basic \SSVax}\label{fig:targeted_one_vax}
    \end{subfigure}
    \begin{subfigure}[b]{.23\textwidth}
        \centering
        \includegraphics[width=\textwidth]{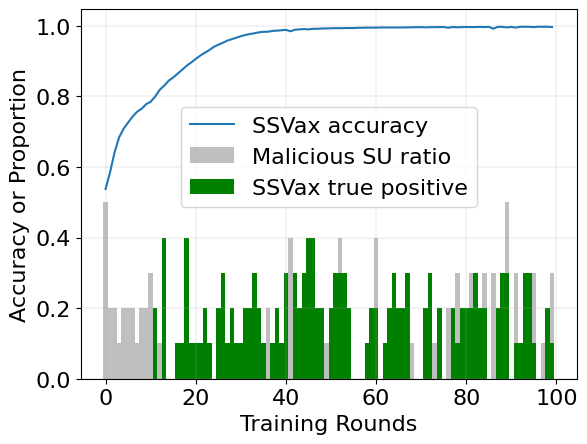}
        \caption{\SSVax\ with LPC}\label{fig:targeted_one_lpc}
    \end{subfigure}
    \begin{subfigure}[b]{.23\textwidth}
        \centering
        \includegraphics[width=\textwidth]{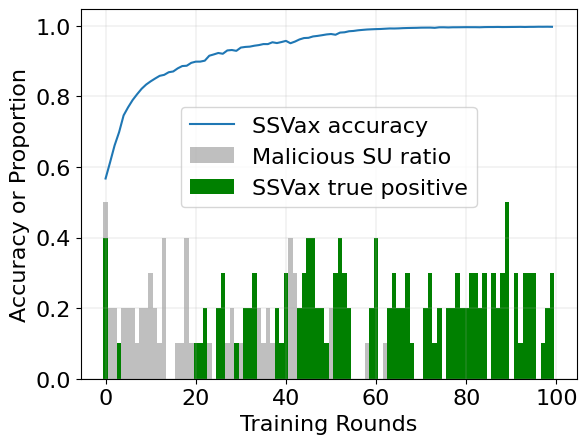}
        \caption{\SSVax\ with APC}\label{fig:targeted_one_apc}
    \end{subfigure}
    \caption{Performance of \SSVax\ against extreme targeted attacks on one channel of the LTE dataset.}
    \label{fig:targeted_one}
\end{figure}

In an experiment targeting a single channel of the LTE dataset, Fig. \ref{fig:targeted_one_vax} shows the true positive rate of \SSVax\ across training rounds. The results demonstrate that \SSVax\ achieves high efficiency as the learning model approaches convergence. During the early rounds, local models exhibit significant divergence, causing \SSVax\ to fail in detecting malicious updates with minimal differences from benign updates (targeted on $1/16$ channels). Similarly, in the later rounds, as the global model stabilizes, \SSVax\ becomes less effective due to the convergence of the learning model and the stealthiness of the extreme targeted attacks. In comparison, the results of \FLTrust\ plotted in Fig. \ref{fig:targeted_one_trust} show that \FLTrust\ faces similar difficulties in defending against these targeted attacks, with high false negative rates in both the early and later rounds. This suggest that classifying benign and malicious updates solely based on the updates themselves becomes more challenging in the extreme cases of targeted attacks, indicating the potential need for additional metrics.

To address this challenge, we adopt the loss impact per class metric used in \FLShield\ to validate updates locally and identify malicious ones. Given the small labeled dataset, the FC can use this metric to validate incoming updates, particularly when attacks target a single channel to evade detection. Instead of classifying updates based solely on their parameters, the model validation results can be used to classify local models, with the expectation that malicious models will show significant deviations in validation results compared to benign ones. We further conduct experiments modifying the basic \SSVax\ to utilize model validation results, instead of model update parameters, as the clustering metric. We validate the models on each channel to address targeted attacks, using both the cross-entropy loss per class (LPC) as described in \cite{FLShield24Kabir} and learning accuracy per class (APC). The results, shown in Figs. \ref{fig:targeted_one_lpc} and \ref{fig:targeted_one_apc}, indicate that these metrics help \SSVax\ effectively detect and filter out malicious updates from targeted attacks. Specifically, in the early rounds, some malicious updates bypass \SSVax\ and are included in aggregation due to the high divergence of local models. However, as learning stabilizes and malicious model updates exhibit distinguishable lower accuracy on targeted channels, abnormal validation results help \SSVax\ identify and filter out malicious updates. Among the metrics, the APC is more effective than LPC in distinguishing between benign and malicious updates, as accuracy is directly related to the attacker's objectives. In other experiments with targeted attacks on two channels, \SSVax\ with APC successfully detects and filters out all malicious updates when the model converges (after $50/100$ rounds). This improved performance comes at the cost of validating model updates at the FC. 

\subsection{Discussion}

\SemiSS\ is designed for SS domain, an increasingly vital research area driven by new network generations and spectrum constraints due to the surge of connected devices. The discovery that processing SS data in the frequency domain outperforms raw IQ data processing, with minimal FFT overhead, offers a significant advantage for mobile applications. This approach enables faster on-device ML, reducing resource demands and minimizing exposure to security threats. 


\SSVax\ relies on a small benign dataset to generate vaccines that effectively filter malicious updates and offering potential explanations for detection decisions. Initially developed for FLSS, this approach is adaptable to broader FL applications where a similar dataset is available at the server, similar to \FLTrust. Although direct comparisons with \FLShield\ and MESAS are not conducted due to conflicting their majority-based assumptions, incorporating metrics from these works enhances \SSVax's performance as previously demonstrated. The modular design of \SSVax\ supports customization of metrics and clustering algorithms, paving the way for future research to optimize its application in diverse use cases.


A key limitation of \SSVax\ is the additional computation cost of vaccine distillation at the FC, which demands extra resources and may delay aggregation. However, this process can run concurrently with SUs training, minimizing delays. Exploring vaccine distillation at trusted SUs could be a direction when an FC dataset is unavailable. Notably, SU operations remain unaffected, with computation and communication costs equivalent to those in undefended FLSS.

Another constraint, inherent to the vaccination approach, is the need to select pseudo-attacks tailored to specific real-world attacks. Future work could investigate more versatile pseudo-attacks capable of countering multiple attack types. Backdoor attacks \cite{BackdoorFL20Bagdasaryan}, a type of targeted attack, are significant in FL but less feasible in FLSS due to difficulty of tampering with input data (sensed signals). However, the potential impact of jammers \cite{SecureFL23Wasilewska} introduces a compelling research avenue.

\section{Conclusion}\label{conclusion}
The growing reliance on mid-band frequencies and the increasing scarcity of wireless spectrum in the 5G and upcoming 6G eras present significant challenges in supporting the vast number of connected devices. As a result, the demand for more reliable and accurate SS solutions has become critical. In this work, we propose \SemiSS\ and \SSVax\ to enhance and secure FLSS for B5G. \SemiSS\ addresses the challenge of training models with unlabeled data, while \SSVax\ offers robust protection against data poisoning attacks by leveraging a vaccine-based approach. Extensive experiments on multiple datasets validate the effectiveness and security of the proposed methods for active SS applications in current and future radio networks. Potential research should focus on evaluating these methods on diverse telecommunication datasets and further optimizing the components of \SSVax\ to improve its adaptability and defense capabilities.

\bibliographystyle{IEEEtran}
\bibliography{refs}

\appendices

\section{Performance of \SemiSS\ under Varying FC Data Distributions}\label{sec:data_distributions}
Table \ref{tab:flss_sdr_acc} presents the learning performance of \SemiSS\ under varying data distributions, where the FC's labeled dataset may not cover all the labels present in the SUs' unlabeled data. The key findings are as follows:

\begin{itemize}
\item For IQ data, the accuracy is highly sensitive to the completeness of the labels in the FC's dataset. When the FC is trained on an incomplete labeled dataset, it struggles to guide the labeling process for unrepresented classes in the SUs' data. This results in a significant drop in accuracy as the number of missing labels increases.
\item For frequency data, remarkably, perfect accuracy is achieved even with a minimal labeled dataset containing only a single class. This highlights the robustness of frequency-domain processing in handling incomplete label distributions and its ability to generalize effectively.
\end{itemize}

\begin{table}[h]\centering\smallsize
\caption{Training performance of \SemiSS\ on the SDR dataset under varying FC data distributions.}\label{tab:flss_sdr_acc}
\begin{tabular}{|l|c|c|c|c|c|}\cline{3-6}
\multicolumn{2}{c|}{} & \multicolumn{4}{c|}{\textbf{Number of class(es)}} \\\hline
\textbf{Experiments on} & $|\mathcal{D}^S|$ & $4/4$ & $4/4$ & $2/4$ & $1/4$ \\\hline\hline
Frequency data & $10$ & $100\%$ & $100\%$ & $100\%$ & $100\%$ \\\hline
Raw IQ data & $200$ & $99.8\%$ & $86.9\%$ & $72.2\%$ & $53.0\%$ \\\hline
\end{tabular}
\end{table}

\newpage

\section{\SSVax\ Defense with Multiple Vaccines Against Label-Poisoning Attacks}\label{sec:multiple_vaccines}
We evaluate \SSVax\ by employing multiple vaccines to defend against diverse label-poisoning attacks. In these experiments, label-flipping and label-setting vaccines are used to counter three attack types: label-flipping, label-setting, and label-random attacks. Label-random attacks simulate scenarios where clients neglect or intentionally mislabel their data, by assigning random labels. Unlike label-flipping or label-setting attacks, which are designed to disrupt the FLSS process intentionally, label-random attacks are less targeted but still undermine the system's objectives. The experimental results demonstrate that \SSVax\ effectively detects and filters out all malicious updates, achieving zero false-negative and false-positive rates. Malicious updates are accurately clustered into groups corresponding to their respective vaccines, with one exception: label-random attack updates are classified under the label-flipping group. This outcome suggests that label-random attacks exhibit behavioral similarities to label-flipping attacks, allowing the corresponding vaccine to defend against them effectively. These results, summarized in Table \ref{tab:clustering}, highlight \SSVax's ability to mitigate various attack types while identifying their specific characteristics. Additionally, we adopt the \FLTrust\ approach to generate a server update and include it in the clustering process, hypothesizing that it would fall into the benign group and enhance clustering confidence. Contrary to expectations, the server update consistently forms a separate cluster. In experiments involving $3$ attack types (label-flipping, label-setting, and label-random), the new cluster includes updates from both label-random attacks and the server. In scenarios with only $2$ attack types (label-flipping and label-setting), the server update still forms an independent cluster. Notably, during the first round of the $3$-attack scenario, one label-random update is initially classified into the server update's cluster. These findings indicate that the server update alone is insufficient to cluster benign and malicious updates effectively, emphasizing the critical role of vaccines in addressing these attack challenges. \SSVax\ not only mitigates current attack types but also provides insight into their nature, reinforcing its robustness in securing FLSS systems.

\begin{table}[h]\centering\smallsize
\caption{Classification results under label-poisoning attacks scenarios.}
\label{tab:clustering}
\begin{tabular}{|c||c|c|c|c|}\hline
\textbf{Attack} & \multicolumn{3}{c|}{\textbf{Updates affected by label-}} & \textbf{Server} \\\cline{2-4}
\textbf{scenario} & \textbf{Flipping} & \textbf{Setting} & \textbf{Random} & \textbf{update} \\\hline\hline
$2$ types & Flipping & Setting & Flipping* & New \\\cline{1-4}
$3$ types & Flipping & Setting & N/A & cluster\\\hline
\multicolumn{5}{l}{*\textit{One case classified as part of the server update cluster.}}
\end{tabular}
\end{table}

\end{document}